%% file: main.tex
\pdfoutput=1
\documentclass{article}

\input{math_commands.tex}

\usepackage{microtype}
\usepackage{graphicx}
\usepackage{subfigure}
\usepackage{booktabs}

\newcommand{\fig}[1]{Figure~\ref{fig:#1}}

\newcommand{\tab}[1]{Table~\ref{tab:#1}}

\usepackage{hyperref}

\usepackage[accepted]{icml2020}

\icmltitlerunning{RPGAN: GANs Interpretability via Random Routing}

\begin{document}

\twocolumn[
\icmltitle{RPGAN: GANs Interpretability via Random Routing}

\begin{icmlauthorlist}
\icmlauthor{Andrey Voynov}{ya}
\icmlauthor{Artem Babenko}{ya,hse}
\end{icmlauthorlist}

\icmlaffiliation{ya}{Yandex, Russia}
\icmlaffiliation{hse}{National Research University Higher School of Economics
, Moscow, Russia}

\icmlcorrespondingauthor{Andrey Voynov}{an.voynov@yandex.ru}

\icmlkeywords{GAN}

\vskip 0.3in
]

\printAffiliationsAndNotice{}

\begin{abstract}
Generative adversarial networks (GANs) are currently the top choice for applications involving image generation.
However, in practice, GANs are mostly used as black-box instruments, and we still lack a complete understanding of an underlying generation process. While several recent works address the interpretability of GANs, the proposed techniques require some form of supervision and cannot be applied for general data.

In this paper, we introduce Random Path Generative Adversarial Network (RPGAN) --- an alternative design of GANs that can serve as a data-agnostic tool for generative model analysis. While the latent space of a typical GAN consists of input vectors, randomly sampled from the standard Gaussian distribution, the latent space of RPGAN consists of random paths in a generator network. As we show, this design allows to understand factors of variation, captured by different generator layers, providing their natural interpretability. With experiments on standard benchmarks, we demonstrate that RPGAN reveals several insights about the roles that different layers play in the image generation process. Aside from interpretability, the RPGAN model also provides competitive generation quality and enables efficient incremental learning on new data. The PyTorch implementation of RPGAN is available online\footnote{\url{https://github.com/anvoynov/RandomPathGAN}}.
\end{abstract}

\input{intro}
\input{related}
\input{method}

\input{experiments}
\input{conclusion}

\bibliography{main}
\bibliographystyle{icml2020}

\appendix
\input{appendix}

\end{document}

%% file: math_commands.tex
\usepackage{amsmath,amsfonts,bm}

\def\eqref#1{equation~\ref{#1}}

\def\1{\bm{1}}

\DeclareMathAlphabet{\mathsfit}{\encodingdefault}{\sfdefault}{m}{sl}
\SetMathAlphabet{\mathsfit}{bold}{\encodingdefault}{\sfdefault}{bx}{n}



%% file: intro.tex
\section{Introduction}
\label{sect:intro}

Nowadays, deep generative models are an active research direction in the machine learning community. The dominant methods for generative modeling, such as Generative Adversarial Networks (GANs), are currently able to produce diverse photorealistic images \cite{big_gan, style_gan}. These methods are not only popular among academicians but are also a crucial component in a wide range of applications, including image editing \cite{isola2017image,zhu2017unpaired}, super-resolution \cite{ledig2017photo}, video generation \cite{wang2018video} and many others.

Along with practical importance, a key benefit of accurate generative models is a more complete understanding of the internal structure of the data. Insights about the data generation process can result both in the development of new machine learning techniques as well as advances in industrial applications. However, most state-of-the-art generative models employ deep multi-layer architectures, which are difficult to interpret or explain. While many works investigate the interpretability of discriminative models \cite{zeiler2014visualizing,simonyan2013deep,mahendran2015understanding}, only a few \cite{chen2016infogan, bau2019gandissect, yang2019semantic} address the understanding of generative ones. Moreover, the techniques proposed in \cite{bau2019gandissect, yang2019semantic} require labeled datasets or pretrained models, which can be expensive to obtain, hence, have limited applicability. Overall, at the moment, there is no tool that provides an understanding of how GANs work on general data, and we aim to reduce this gap by our paper.

In this work, we propose the Random Path GAN (RPGAN) --- an alternative design of GANs that allows the natural interpretability of the generator network. In traditional GAN generators, the stochastic component that influences individual samples is a noisy input vector, typically sampled from the standard Gaussian distribution. In contrast, RPGAN generators instead use stochastic routing during the forward pass as their source of stochasticity. In a nutshell, the RPGAN generator contains several instances of each generator layer, and only one random instance is activated during generation. The training of the RPGAN can is performed in the same adversarial manner as in traditional GANs. In the sections below, we show how RPGAN allows to understand the factors of variation captured by the particular layer and reveals several interesting findings about the image generation process, e.g., that different layers are ``responsible for'' coloring or objection location. As a practical advantage, RPGANs can be efficiently updated to new data via the simple addition of new instances of a particular layer, avoiding re-training the full model from scratch. Finally, we observe that RPGANs allow to construct generative models without nonlinearities, which can significantly speed up the generation process for fully-connected layers. In summary, the main contributions of our paper are:

\begin{figure*}
\begin{tabular}{cc}
\includegraphics[width=0.94\columnwidth]{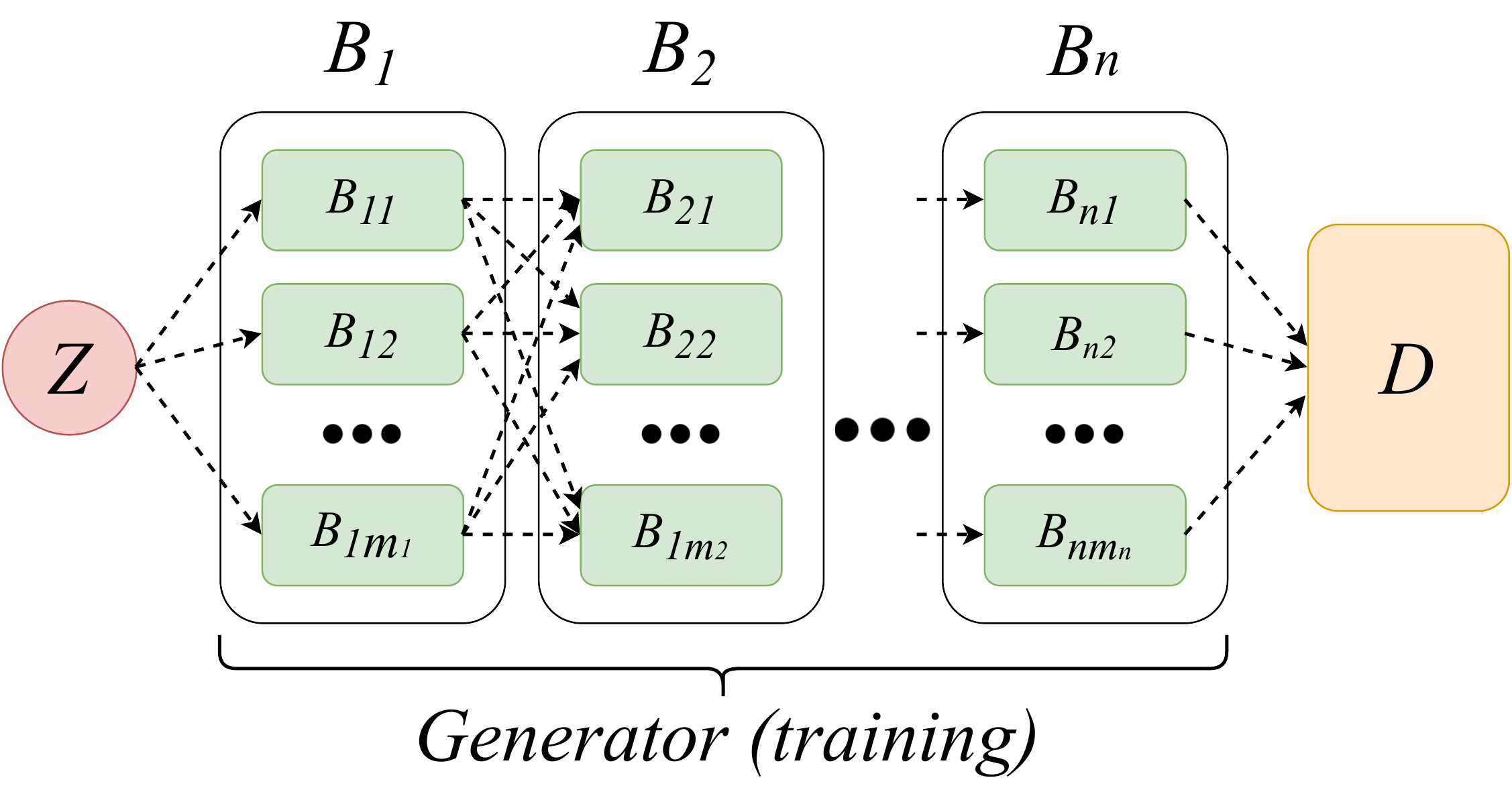} &
\includegraphics[width=0.97\columnwidth]{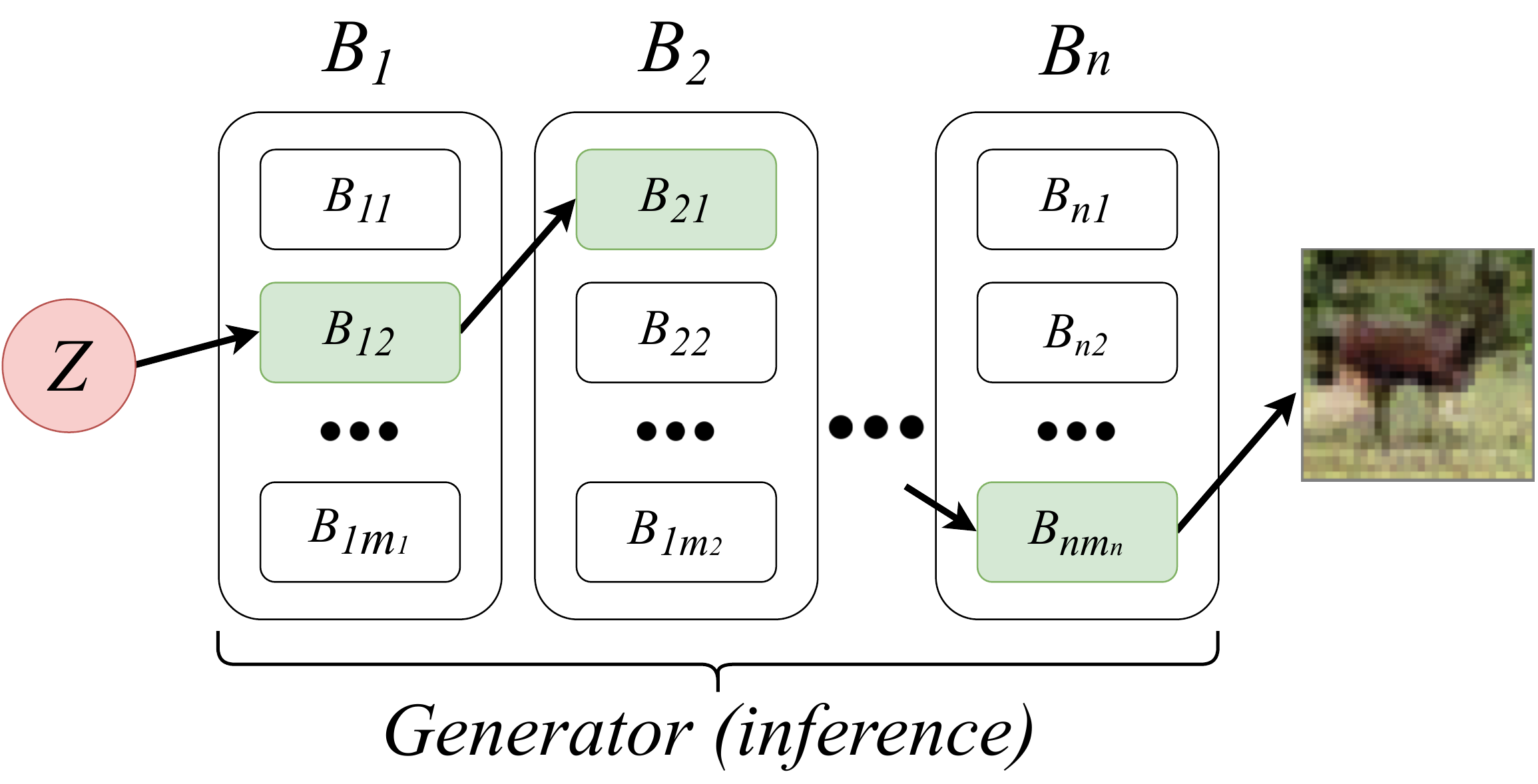}
\end{tabular}
\vspace{-4mm}
\caption{Design of the proposed RPGAN model. The RPGAN generator consists of several buckets $B_1,\dots,B_n$, corresponding to layers of the generator network. Each bucket contains several instances of the corresponding layer. During forward pass, a random instance from each bucket is ``activated'' to generate a particular image.}
\label{fig:rpgan_model}
\vspace{-3mm}
\end{figure*}


\begin{itemize}
    \item We introduce RPGAN --- GAN with an alternative source of stochasticity, based on random routing, enabling to analyse the roles of different layers in the generation process. To the best of our knowledge, RPGAN is the first \textbf{completely unsupervised} technique for GANs interpretability.
    
    \item With experiments on standard benchmarks, we reveal several insights about the generation process. Many insights confirm and extend the findings from prior works \cite{bau2019gandissect, yang2019semantic} that exploit some form of supervision. Note RPGAN is more general compared to the techniques from \cite{bau2019gandissect, yang2019semantic} as RPGAN does not require labeled data or pretrained models.
    
\end{itemize}

%% file: related.tex
\section{Related work}
\label{sect:related}

In this section we briefly describe connections of RPGAN to existing ideas from prior works

\textbf{Generative adversarial networks.} GANs are currently one of the main paradigms in generative modelling. Since the seminal paper on GANs by \cite{goodfellow2014generative}, a plethora of alternative loss functions, architectures, normalizations, and regularization techniques were developed \cite{kurach2019large}. Today, state-of-the-art GANs are able to produce high-fidelity images, often indistinguishable from real ones \cite{big_gan, style_gan}. In essence, GANs consist of two networks -- a generator and a discriminator, which are trained jointly in an adversarial manner. In standard GANs, the generation stochasticity is provided by the input noise vector. In RPGANs, we propose an alternative source of stochasticity by using a fixed input but random routes during forward pass in the generator. 

\textbf{Specific GAN architectures.} Many prior works investigated different design choices for GANs, but to the best of our knowledge, none of them explicitly aimed to propose an interpretable GAN model. \cite{hoang2018mgan} proposed the use of several independent generators to address the mode collapse problem. \cite{Chavdarova_2018_CVPR} employ several auxiliary local generators and discriminators to improve mode coverage as well. \cite{Huang_2017_CVPR} use layer-wise generators and discriminators to enforce hidden representations produced by layers of the generator to be similar to the corresponding representations produced by a reversed classification network. Important differences of RPGAN compared to the works described above is that it uses random routes as its latent space and does not enforce to mimic the latent representations of pretrained classifiers.

\textbf{Interpretability.} While the interpretability of models based on deep neural networks is an important research direction, most existing work addresses the interpretability of discriminative models. These works typically aim to understand the internal representations of networks \cite{zeiler2014visualizing, simonyan2013deep, mahendran2015understanding,  dosovitskiy2016generating} or explain decisions produced by the network for particular samples \cite{sundararajan2017axiomatic,bach2015pixel,simonyan2013deep}. However, only a few works address the interpretability of generative models. Related work by \cite{bau2019gandissect} develops a technique that allows to identify which parts of the generator are responsible for the generation of different objects. Note, that the technique from \cite{bau2019gandissect} requires a pretrained segmentation network and cannot be directly applied to several benchmarks, e.g., CIFAR-10 or MNIST. A recent work \cite{yang2019semantic} also aims to understand the roles of different generator layers, but their approach relies on pretrained classifiers; hence, it has limited applicability.

In contrast, RPGAN does not require any auxiliary models or other forms of supervision and can be applied to general data.  Some of our findings confirm the results from \cite{bau2019gandissect, yang2019semantic}, providing stronger evidence about the roles of different layers in the generation process. 

%% file: method.tex
\section{Random Path GAN}
\label{sect:method}

\subsection{Motivation}

Before a formal description, we provide intuition behind the RPGAN model. Several prior works have demonstrated that in discriminative convolutional neural networks, different layers are ``responsible'' for different levels of abstraction \cite{zeiler2014visualizing, babenko2014neural}. For instance, earlier layers detect small texture patterns, while activations in deeper layers correspond to semantically meaningful concepts. Similarly, in our paper we aim to understand the roles that different GAN layers play in image generation. Thus, we propose an architecture that provides a direct way to interpret the impact of individual layers. For a given architecture, RPGAN has several instances of each layer in its architecture. During the forward pass, a random instance of each layer is used to generate a particular image. Therefore, one can analyze the role of RPGAN layers by visualizing how different instances of each layer affect the image.

\subsection{Model}
Here we formally describe a structure of the RPGAN model.
Similarly to the standard GAN architectures, RPGAN consists of two networks -- a generator and a discriminator. The RPGAN discriminator operates exactly like discriminators in standard GANs, so we focus on the generator description.

The RPGAN generator consists of several consequent \textit{buckets} $B_1, \dots, B_n$, where each bucket corresponds to a particular computational unit of a generator, e.g., a ResNet block \cite{He2015DeepRL}, a convolutional layer with a nonlinearity or any other. We associate each bucket with a layer (or several layers) in the generator architecture, which we aim to interpret or analyze. In each bucket, RPGAN maintains several independent instances of the corresponding computational unit $B_i{=}\{B_{i1}, \dots, B_{im_i}\}$, see \fig{rpgan_model}.

RPGAN generator performs forward pass as follows. For each $i{=}1, \dots, n - 1$ a random instance from the bucket $B_i$ produces an intermediate output tensor that is passed to a random instance from the next bucket $B_{i + 1}$. 
An instance from the first bucket $B_1$ always receives a fixed input vector $Z$, which is the same for different forward passes. Therefore, the generator stochasticity arises only from a random path that goes from $Z$ to an output image, using only a single instance from each bucket. Formally, during each forward pass, we randomly choose indices $s_1, \dots, s_n$ with $1 \leq s_i \leq m_i$. The generator output is then computed as $B_{ns_n} \circ \cdots B_{2s_2} \circ B_{1s_1}(Z)$, see \fig{rpgan_model}. In other words, the generator defines a map from the Cartesian product $\langle m_1 \rangle \times \langle m_2 \rangle \times \cdots \times \langle m_n \rangle$ to the image space. Note that we can take an arbitrary existing GAN model, group its generator layers into buckets and replicate them into multiple instances. In these terms, the original model can be treated as RPGAN with a single instance in each bucket and random input noise. Note that during generation, RPGAN performs the same number of operations as standard GANs.

By its design, RPGAN with buckets $B_1, \dots, B_n$ and a constant input $Z$ is able to generate at most $|B_1| \times \cdots \times |B_n|$ different samples were $|B_k|$ is the number of instances in the bucket $B_k$. Nevertheless, this number is typically much larger compared to the training set size. We argue that the probability space of random paths can serve as a latent space to generate high-quality images, as confirmed by the experiments below.  The model is highly flexible to the choice of generator and discriminator architectures as well as to the loss function and learning strategy.

\textbf{Instance diversity loss.} To guarantee that the instances in a particular bucket are different, we also add a specific diversity term in the generator loss function. The motivation for this term is to prevent instances $B_{ki}, B_{kj}$ from learning the same weights.
Let $W$ be the set of all parameters of the generator. For each parameter $w \in W$ there is a set of its instances $\{w^{(1)}, \dots w^{(m_w)}\}$ in the RPGAN model. Then we enforce the instances to be different by the additional loss term $-\sum\limits_{w \in W, i \neq j} \text{MSE}\left(\frac{w^{(i)}}{s_w}, \frac{w^{(j)}}{s_w}\right)$. Here we also normalize by the standard deviation of $s_w$ of all parameters from different instances that correspond to the same layer. This normalization 
effectively guarantees that all buckets contribute to the diversity term.

%% file: experiments.tex
\section{Experiments}
\label{sect:experiments}

\begin{table}[!b]
\addtolength{\tabcolsep}{-4pt}
\centering
\vspace{-5mm}
    \begin{tabular}{ |c|c|c|c|c|c|c| }
        \hline
        Dataset & Size & Buckets & $n_{in}$ & $d_{steps}$ & Batch & Coverage\\
        \hline
        CIFAR-10 & $32{\times32}$ & $5$ & $40$ & $5$ & $64$ & $2048$ \\ 
        AnimeFaces & $64{\times}64$ & $6$ & $20$ & $1$ & $32$ & $\approx 2970$ \\
        LSUN & $128{\times}128$ & $7$ & $20$ & $1$ & $16$ & $\approx 420$ \\ 
        \hline
    \end{tabular}
\vspace{-3mm}
\caption{The details of architectures and training protocols.}
\label{tab:dataset_hypperparams}
\end{table}

\textbf{Architecture.} In all the experiments, we use ResNet-like generators with spectral normalization and the hinge loss (SN-ResNet) as in \cite{miyato2018spectral}. The instances in the first bucket are fully-connected layers, the instances in the last bucket are convolutional layers and instances in all other buckets are residual blocks with two convolutions and a skip connection. If not stated otherwise, all the buckets have the same number of instances. Additional experiments with other architectures are provided in supplementary.

\textbf{Datasets.} We performed experiments on CIFAR-10 \cite{CIFAR10}, LSUN-bedroom \cite{LSUN} and Anime Faces \cite{AnimeFaces} datasets. For different datasets, we use different numbers of discriminator steps per one generator step $d_{steps}$ and different numbers of instances in each bucket $n_{in}$. The parameters used for three datasets are summarized in \tab{dataset_hypperparams}. In the last column, we also report \textit{Coverage}, which is the ratio of the latent space cardinality (which equals the number of buckets to the power $n_{in}$) to the dataset size. Intuitively, large coverage guarantees that RPGAN has a sufficiently rich latent space of generator routes to capture the reference dataset. In the experiments below, we demonstrate that even moderate coverage is sufficient to generate high-fidelity images (see the LSUN-bedroom dataset with coverage $\approx 420$).

\textbf{Training details.} We use the Adam optimizer with learning rate equal to $0.25 \times 10^{-3}$, $\beta_1, \beta_2$ equal to $0.5, 0.999$ and train the model for $45 \times 10^4$ generator steps for CIFAR-10 and $25 \times 10^4$ generator steps for Anime Faces and LSUN-bedroom datasets. During training, we also learn the unique input vector $Z$. We observed that a learnable $Z$ slightly improves the final generation quality and stabilizes the learning process. Training is performed by passing $Z$ through $N$ independent random paths. Formally, let $\{x_1, \dots, x_N\}$ be a batch of samples received from a bucket $B_k$. To pass this batch through the bucket $B_{k + 1}$ we take random instances $B_{ki_1}, \dots, B_{ki_N}$ and form a new batch $\{B_{ki_1}(x_1), \dots, B_{ki_N}(x_N)\}$. In all the experiments, we use the same training protocols for both RPGAN and the standard GAN of the same generator architecture. Note that despite a larger number of learnable parameters, RPGAN does not require more data or training time to achieve the same quality, compared to standard GANs.

\input{experiments_subsections/layers_specification}
\input{experiments_subsections/relation}
\input{experiments_subsections/blocks_count_ablation}
\input{experiments_subsections/dataset_extension}
\input{experiments_subsections/pure_linear_generator}

%% file: experiments_subsections/layers_specification.tex
\subsection{Do different generator layers affect different factors of variations?}
\label{sect:layer_specification}

\begin{figure}
    \centering
    \includegraphics[width=0.375\paperwidth]{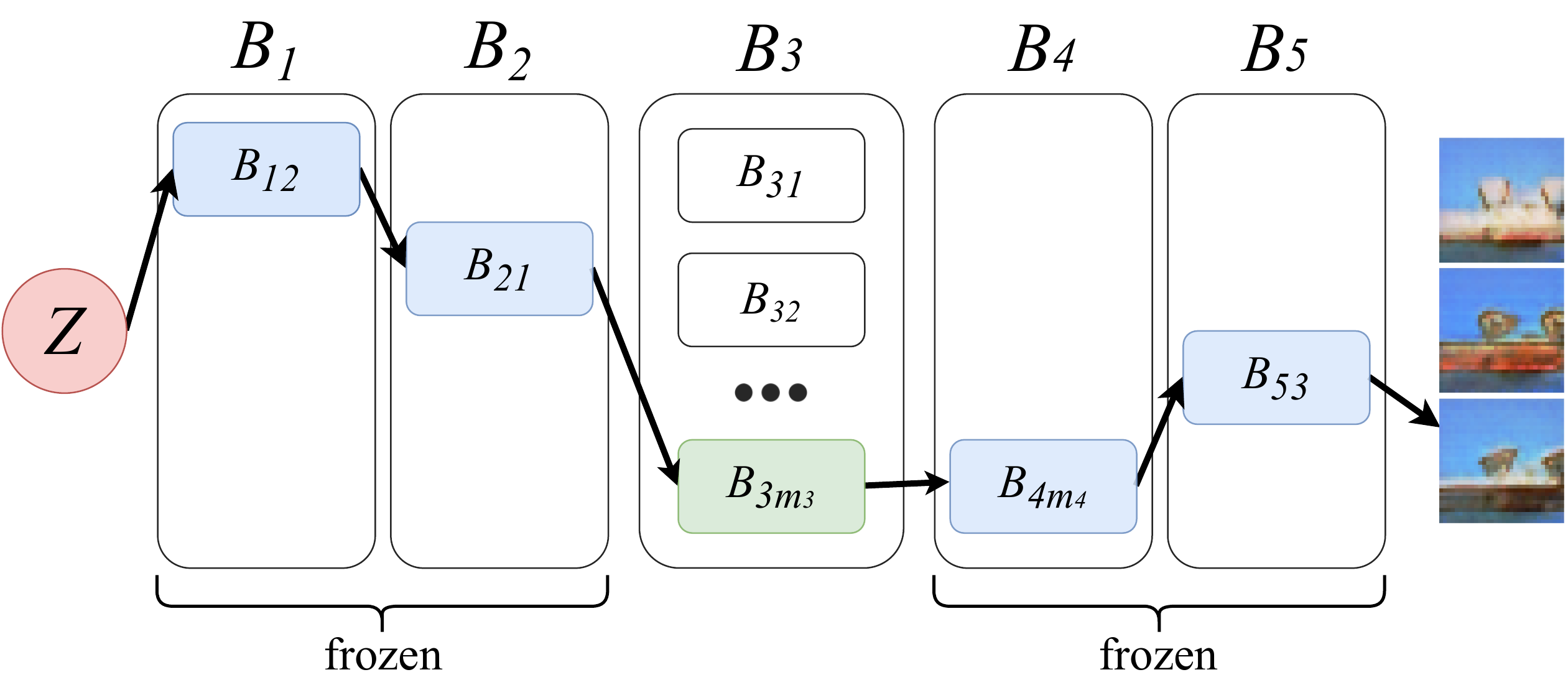}
    \vspace{-3mm}
    \caption{The five-bucket RPGAN model with ``frozen'' instances in buckets 1, 2, 4 and 5. In this case the stochasticity comes only from the third bucket. Analyzing the generated images allows to identify factors of variation, affected by the third bucket.}
    \vspace{-5mm}
    \label{fig:rpgan_layer_freeze}
\end{figure}

In the first series of experiments, we investigate the ``responsibility'' of different generator layers. With RPGAN, this can be performed with a technique schematically presented on \fig{rpgan_layer_freeze}. In this example, the goal is to understand the role of the third bucket $B_3$ in a five-bucket generator. To this end, we fix the instances from all other buckets $B_1,B_2,B_4,B_5$, shown in blue on \fig{rpgan_layer_freeze}. Then we generate images corresponding to routes that contain all the fixed instances, with the stochasticity coming only from varying instances from the target bucket $B_3$. By inspecting the distribution of the obtained images, one can understand what factors of variation are affected by $B_3$.

\begin{figure}
    \centering
    \includegraphics[width=0.375\paperwidth]{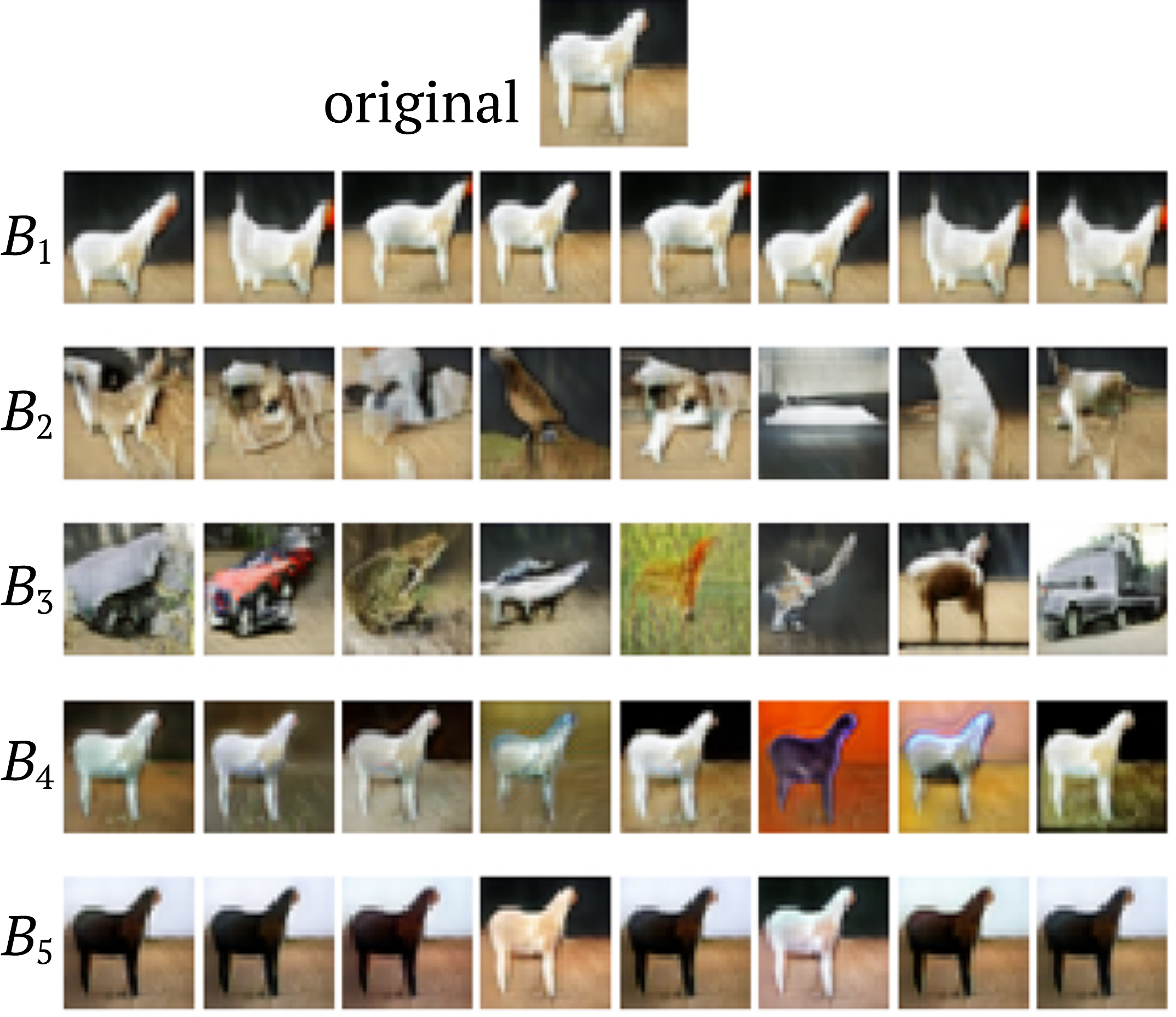}
    \caption{\textit{Top image}: an image generated with a fixed sequence of instances. \textit{Horizontal lines}: images generated by the same sequence of instances in all buckets and one \textit{unfrozen} bucket. In the unfrozen bucket we choose eight arbitrary instances to avoid excessively large figures. The generated images allow to interpret factors of variation, captured by different buckets.} 
    \vspace{-5mm}
    \label{fig:rpgan_layer_freeze_out}
\end{figure}

\fig{rpgan_layer_freeze_out} shows an example of image distributions obtained by varying instances in different buckets of five-bucket RPGAN for CIFAR-10. Each row shows how the original generated image can change if different instances from the corresponding bucket are used during generation. Other qualitative examples for different datasets are provided in the supplementary material. Several observations from these figures are the following. The first bucket typically does not influence coloring and mostly affects small objects' deformations. The intermediate buckets have the largest influence on semantics. The last two buckets are mostly responsible for coloring and do not influence the content shape. In particular, on \fig{rpgan_layer_freeze_out}, the fourth layer widely varies color, while the fifth acts as a general tone corrector. Note that these observations are consistent with the insights revealed by \cite{bau2019gandissect,yang2019semantic} but are obtained without any supervision. In contrast, techniques from \cite{bau2019gandissect,yang2019semantic} often cannot be applied since they require pretrained segmentation or classification models (which can be unavailable, e.g., for CIFAR and AnimeFaces).


To verify the observations above, we perform more rigorous experiments that evaluate the roles of different layers quantitatively. Let us define a metric $d_\textbf{img}$ that evaluates the similarity between two generated images. Note that different metrics are able to capture different factors of variations (e.g., in terms of semantic, color histogram, etc.), and we describe two particular choices of $d_\textbf{img}$ below. Then we choose a random route in the RPGAN generator and for each bucket $B_l$ generate four images $Im_1^{(l)}, \dots, Im_4^{(l)}$, varying instances in $B_l$. In other words, we take four random images from each line of the table in the Figure \ref{fig:rpgan_layer_freeze_out}. Then we measure diversity w.r.t. $d_\textbf{img}$ captured by $B_l$ as a ratio 

\begin{equation}
\vspace{-3mm}
D_{l \to 1, d_\textbf{img}} = \frac{\sum\limits_{i \neq j}d_\textbf{img}(Im_i^{(l)}, Im_j^{(l)})}{\sum\limits_{i \neq j}d_\textbf{img}(Im_i^{(1)}, Im_j^{(1)})}
\vspace{-1.5mm}
\end{equation}

Intuitively, the formula above computes the relative diversity with respect to the first bucket, which typically captures the smallest amount of variations in our experiments. We then average these ratios over 100 independent routes. In essense, higher values of averaged ratio $D_{l \to 1, d_\textbf{img}}$ imply higher diversity of $B_l$ compared to the first bucket in terms of the metric $d_\textbf{img}$, which implies that $B_l$ strongly affects the factor of variation captured by particular metric $d_\textbf{img}$.

\begin{figure}
    \centering
    \includegraphics[width=1.0\columnwidth]{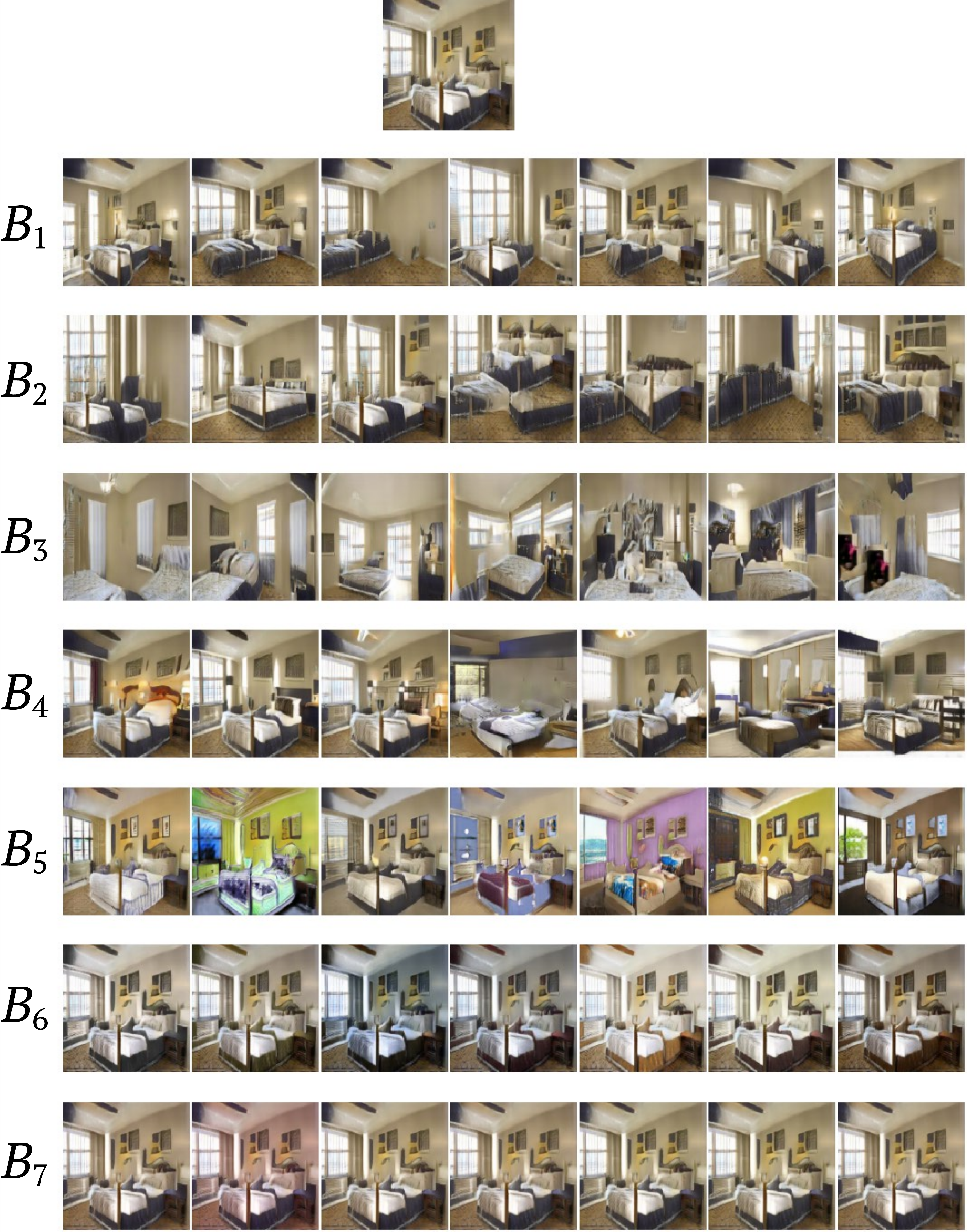}
    \caption{Image distributions for 7-bucket RPGAN and LSUN.}
    \vspace{-5mm}
    \label{fig:layers_variation_lsun}
\end{figure}

\begin{figure}[!b]
    \centering
    \vspace{-3mm}
    \includegraphics[width=1.0\columnwidth]{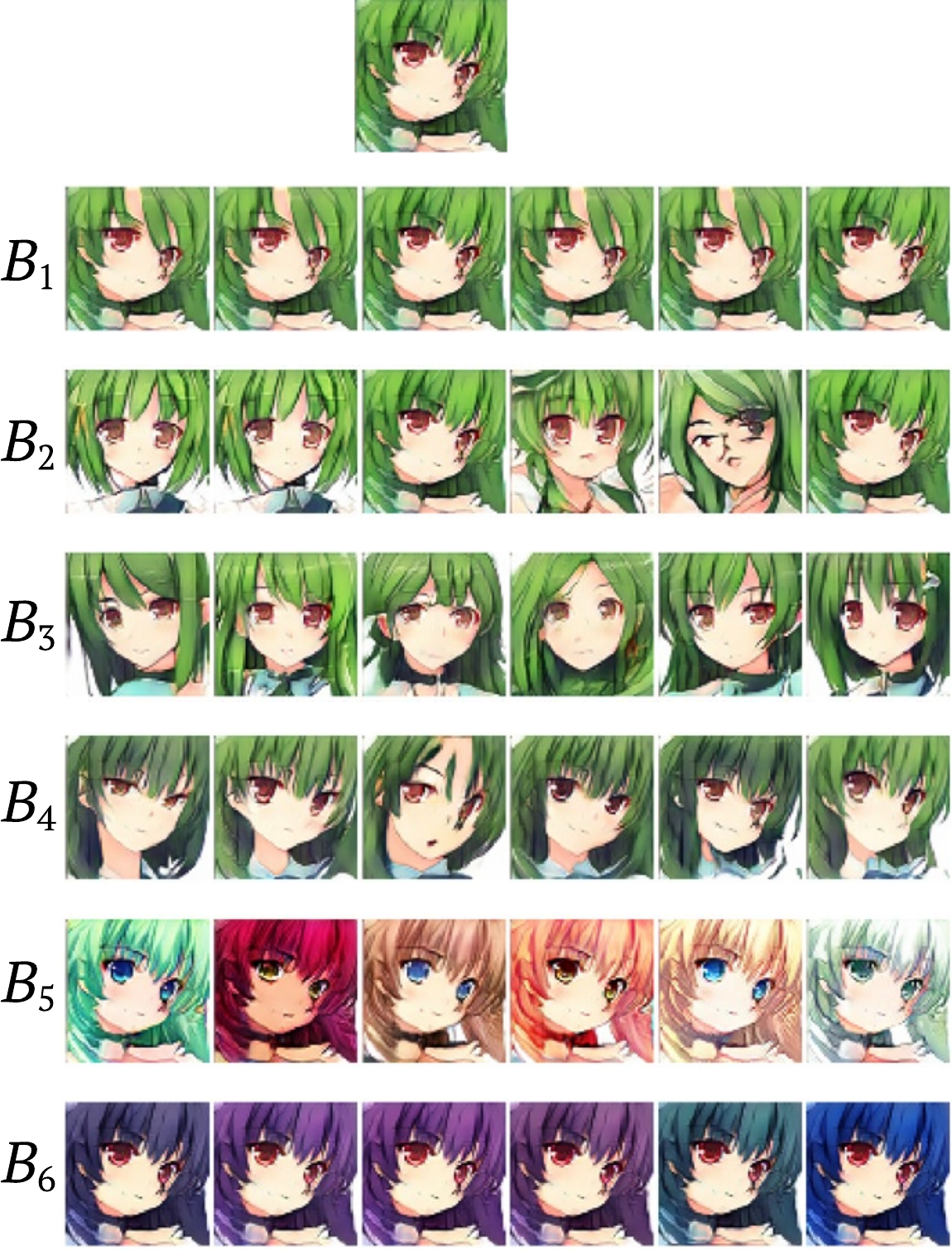}
    \vspace{-7mm}
    \caption{Image distributions for 6-bucket RPGAN and AnimeFaces.}
    \label{fig:layers_variation_anime}
\end{figure}

In experiments, we use two following metrics:

\begin{itemize}
    \item $d_\textbf{semantic}(img_1, img_2)$, capturing semantic, is based on the recent LPIPS \cite{zhang2018unreasonable}, which was shown to indicate perceptual similarity.
    \item $d_\textbf{color}(img_1, img_2)$, measuring difference in colorings. Namely, we take the Hellinger distance between color histograms of generated samples. For each color channel, we split the range $[0, \dots, 255]$ into $25$ equal buckets and evaluate the discrete distribution defined by the frequencies the image pixel intensities appear in a given bucket. Then the Hellinger distance between two quantified color distributions is defined as $d_\textbf{color}(img_1, img_2) = \frac{1}{\sqrt{2}}\sqrt{\sum\limits_{i = 1}^{25}(\sqrt{p_i} - \sqrt{q_i})^2}$ . We compute it for each RGB channel independently.
\end{itemize}

The plots of averaged values of both metrics on three datasets are presented on \fig{rpgan_layer_diversity_cifar}, \fig{layers_variation_anime} \fig{layers_variation_lsun}. They reaffirm that the semantic diversity is the largest for the intermediate layers. On the contrary, the last buckets, which are closer to the output, do not influence semantics but have a higher impact in terms of color. Note that the first layer always shows the smallest variability in terms of both factors of variation. The last bucket affects color correction and color inversion and has a lower pallet variability impact. Overall, we summarize the findings that are common for CIFAR-10, LSUN, and Anime Faces datasets as:

\begin{figure*}[!t]
        \includegraphics[width=0.99\textwidth]{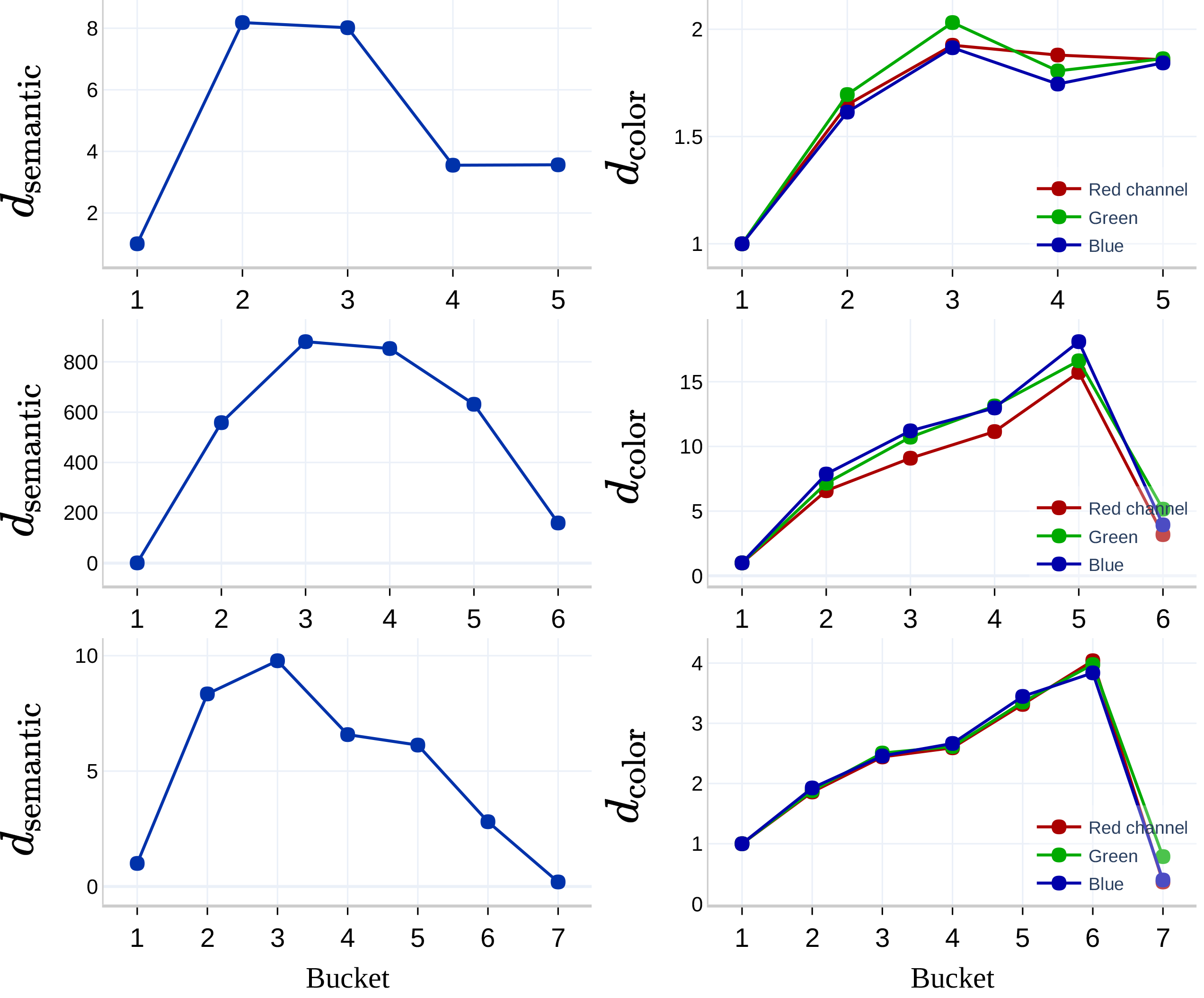}
    \vspace{-4mm}
\caption{The quantitative evaluation of the extent $D_{l \to 1, d}$ to which different generator parts affect different factors of variation for (Top) the five-bucket RPGAN and CIFAR-10; (Middle) the six-bucket RPGAN and AnimeFaces; (Bottom) the seven-bucket RPGAN and LSUN;}
\vspace{-4mm}
\label{fig:rpgan_layer_diversity_cifar}
\end{figure*}

\begin{itemize}
    \item The earlier layers have a smaller variability and seem to be responsible for the viewpoint and the position of the object on the image.
    \item The semantic details of the image content are mostly influenced by the intermediate layers.
    \item The last layers typically affect only coloring scheme and do not affect content semantics or image geometry.
\end{itemize}

Note, that these conclusions can differ for other datasets or other generator architectures. For instance, for the four-bucket generator and MNIST (\fig{mnist_extension}, left) or randomly colored MNIST (~\fig{colred_mnist_variation}, left) the semantics are mostly determined by the first two buckets. Overall, the layers' ``responsibilities'' depend on both an architecture and a particular dataset. Unlike prior techniques, RPGAN can be applied to any generator model and data domain, hence, provides a universal instrument for GAN interpretability.

%% file: experiments_subsections/relation.tex
\subsection{Are RPGAN interpretations valid for standard GANs?}
\label{sect:relation}

In this subsection, we argue that the interpretations of different layers, obtained with RPGAN, are also valid for a standard GAN generator of the same architecture.
First, we demonstrate that both standard GAN and RPGAN trained under the same training protocol provide almost the same generation quality. As a standard evaluation measure,  we use the Fr\'{e}chet Inception Distance (FID) introduced in \cite{heusel2017gans}. 
For evaluation on CIFAR-10, we use 50000 generated samples and the whole train dataset. For both standard GAN and RPGAN, we use ten independently trained generators and report minimal and average FID values in \tab{fid}. In terms of FID RPGAN and the standard GAN perform with comparable generation quality. 

\begin{table}[!htb]
    \centering
    \vspace{-2mm}
    \begin{tabular}{ |c|c|c| }
        \hline
        model & min FID & average FID\\
        \hline
        Five-bucket RPGAN & $16.9$ & $20.8$\\
        SN-ResNet & $16.75$ & $18.7$\\
        \hline
    \end{tabular}
    \vspace{-2mm}
\caption{FID values for CIFAR-10.}
\label{tab:fid}
\end{table}

\begin{figure}[!t]
    \centering
    \includegraphics[width=0.99\columnwidth]{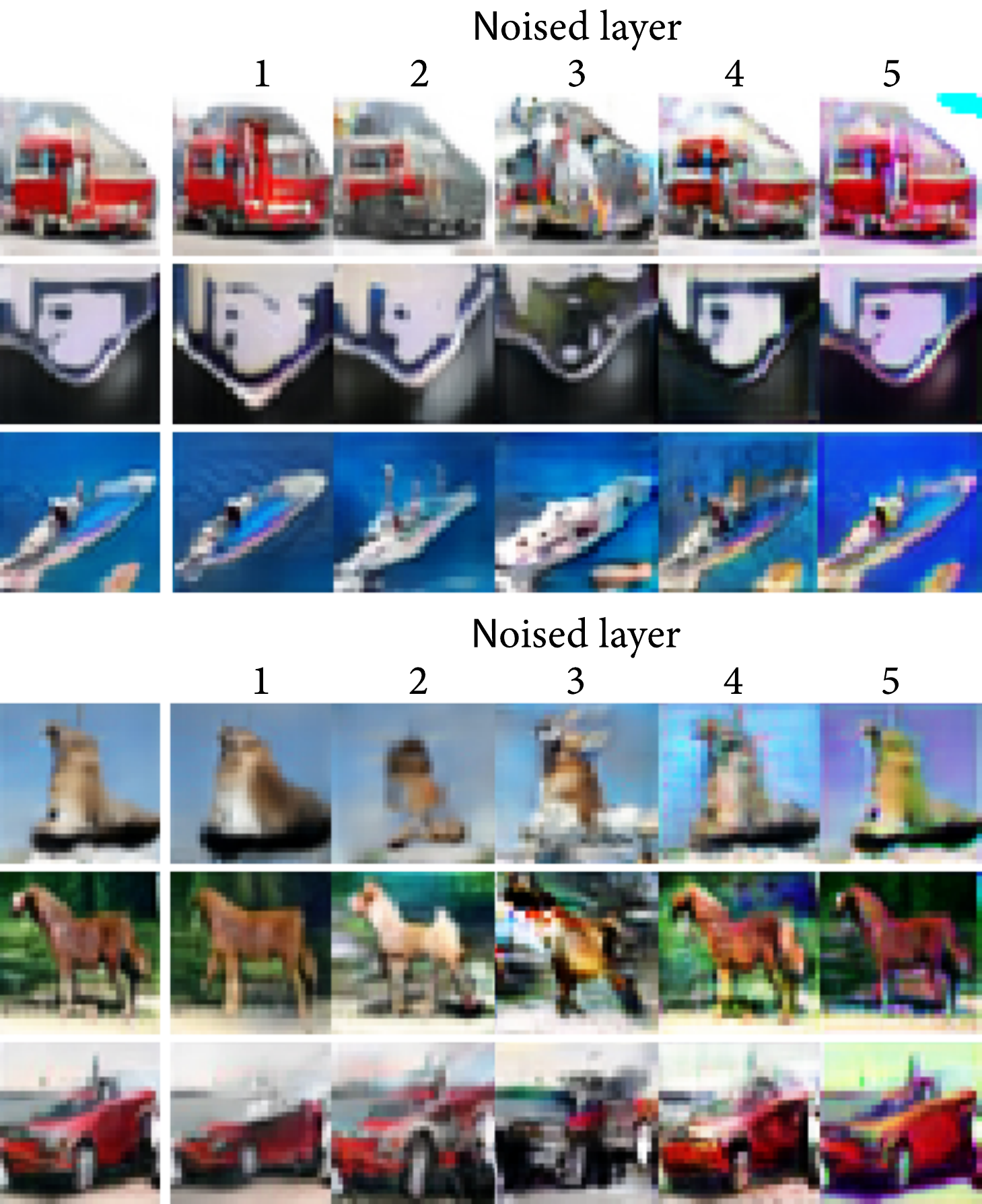}
    \caption{Images, produced by the standard SN-ResNet GAN with noise injection in the parameters of different generator layers. First column: original images, produced without generator perturbation. Other columns: images produced by perturbed generator.}
    \label{fig:noised_vanila}
\end{figure}

\begin{figure}
    \centering
    \includegraphics[width=1.0\columnwidth]{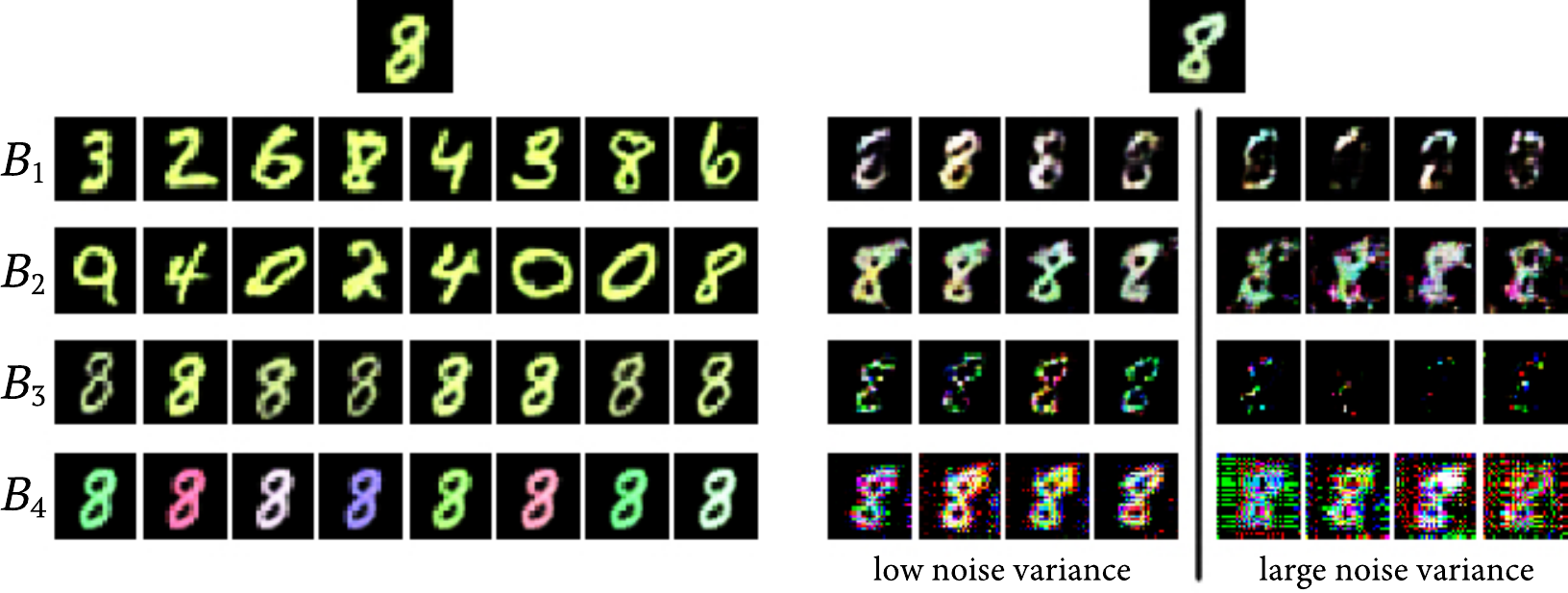}
    \vspace{-3mm}
    \caption{\textit{Left}: images produced by varying instances in a particular bucket of RPGAN. \textit{Right}: images produced by the standard GAN after parameters perturbation in a particular generator layer, with low and high normal noise variance.}
    \vspace{-2mm}
    \label{fig:colred_mnist_variation}
\end{figure}

To confirm that the layers of the standard GAN generator can be interpreted in the same way as the corresponding layers of its RPGAN counterpart, we perform the following experiment. We take a standard SN-ResNet GAN, consisting of five layers associated with the correspondent buckets in RPGAN, and train it on CIFAR-10. Then for each layer, we add normal noise to its weights. Intuitively, we expect that the noise injection in the particular layer will change generated images in terms of factors of variation, influenced by this layer. For instance, noise in the last two layers is expected to harm the coloring scheme, while noise in the intermediate layers is expected to bring maximal semantic damage. Several images, produced by perturbed layers, are presented on \fig{noised_vanila}. The images support the intuition described above and confirm that RPGAN may serve as an analysis tool for the underlying generator model. Note, however, that injecting noise per se cannot be used as a stand-alone interpretability method. The perturbed generators produce poor images, which are difficult to analyze. Meanwhile, RPGAN always generates good-looking images, which allows to identify the factors of variation, corresponding to the particular layer. For instance, see \fig{colred_mnist_variation} for the colored MNIST dataset. \fig{colred_mnist_variation} (left) shows plausible images, generated by varying RPGAN instances. In contrast, \fig{colred_mnist_variation} (right) demonstrates images from generators perturbed with small and large noise. For both noise magnitudes, these images are difficult to interpret.
Of course, given the interpretations obtained via RPGAN, one can perceive similar patterns in the noisy generations, but noise injection alone is not sufficient for interpretability.

%% file: experiments_subsections/blocks_count_ablation.tex
\subsection{Ablation on number of instances}
\label{sect:bloks_count_ablation}
Here we investigate the impact of $n_{in}$ on the generation quality. We train RPGAN with the SN-ResNet generator on CIFAR-10 with a different values of $n_{in}$. The resulting FID values are presented on \tab{n_block_variation}.

\begin{table}[!h]
\addtolength{\tabcolsep}{-4pt}
\centering
    \begin{tabular}{|c|c|c|c|c|c|c|c|c|c|c|}
        \hline
        $n_{in}$ & $5$ & $10$ & $15$ & $20$ & $25$ & $30$ & $35$ & $40$ & $45$ & $50$ \\
        \hline
        FID & $61.0$ & $25.6$ & $23.4$ & $20.0$ & $19.7$ & $18.0$ & $17.4$ & $17.1$ & $18.6$ & $22.9$\\
        \hline
    \end{tabular}
    \vspace{-3mm}
\caption{FID values for RPGAN with different $n_{in}$.}
\label{tab:n_block_variation}
\end{table}

Overall, if $n_{in}$ is too low, the latent space has insufficient cardinality to model real data. On the other hand, large $n_{in}$ results in difficult training and can fail.

%% file: experiments_subsections/dataset_extension.tex
\subsection{Incremental learning with RPGAN}
\label{sect:dataset_extension}

\begin{figure}
    \centering
    \includegraphics[width=0.99\columnwidth]{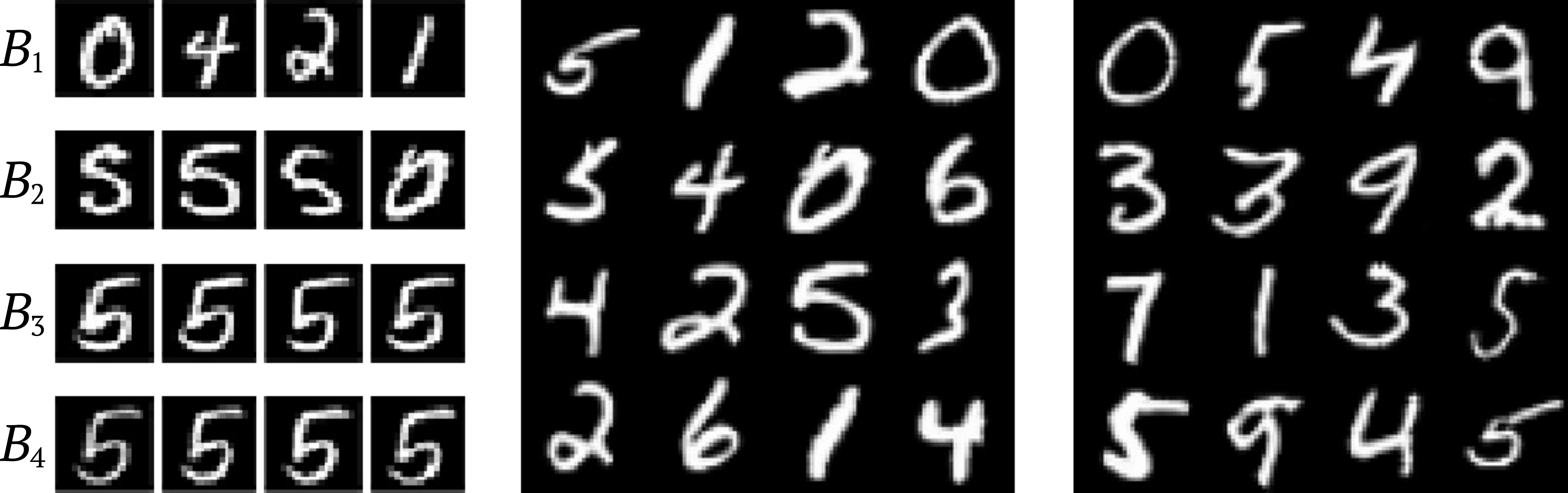}
    \vspace{-3mm}
    \caption{Incremental learning with RPGAN. \textit{Left:} variations captured by different buckets with generator trained on $\text{MNIST}_{0-6}$. \textit{Center:} images produced by a generator trained on $\text{MNIST}_{0-6}$. \textit{Right}: images produced by a generator tuned on $\text{MNIST}$ with only ten new instances training (see the details in the main text).}
    \vspace{-3mm}
    \label{fig:mnist_extension}
\end{figure}

In the next experiment, we demonstrate that the RPGAN model is also a natural fit for the generative incremental learning task (see, e.g., \cite{wu2018memory}). Let us assume that the whole train dataset $D$ is split into two disjoint subsets $D = D_1 \cup D_2$. Suppose that originally we have no samples from $D_2$ and train a generative model to approximate a distribution defined by the subset $D_1$. Then, given additional samples from $D_2$, we aim to solve an incremental learning task --- to update the model with new data without re-training it from scratch. The RPGAN model naturally allows to solve this task efficiently. First, we train an RPGAN generator with buckets $B_1, \dots, B_n$ to approximate the distribution $D_1$. Once one aims to extend the generator with samples from $D_2$, one can add several new instances to the buckets that are responsible for the features that capture the difference between $D_1$ and $D_2$. Then we optimize the generator to reproduce both $D_1$ and $D_2$ by training only the new instances. Thus, instead of training a new generator from scratch, we exploit the pretrained instances that are responsible for features, which are common for $D_1$ and $D_2$. To illustrate this scenario, we take a partition of the MNIST hand-written digits dataset \cite{MNIST} into two subsets $\text{MNIST}_{0-6}$ and $\text{MNIST}_{7-9}$ of digits from $0$ to $6$ and from $7$ to $9$ correspondingly. As for generator for $\text{MNIST}_{0-6}$ we take a four-bucket RPGAN model with $n_{in}$ equals to 20, 20, 20, 8. Note that the last bucket is much thinner than the others, as it turns out to be responsible for variations in writing style, which does not change much across the dataset. Then we train the generator on the subset $\text{MNIST}_{0-6}$ of first 7 digits (see \fig{mnist_extension}, left and center). After that, we add five additional instances to each of the first two layers, obtaining a generator with $n_{in}$ equals to 25, 25, 20, 8, and pretrained weights in all instances except for five in the first and in the second buckets. Then we train the extended model to fit the whole MNIST by optimizing only the ten new instances (see \fig{mnist_extension}, right).

%% file: experiments_subsections/pure_linear_generator.tex
\begin{table}[!b]
\centering
\renewcommand{\arraystretch}{1.2}
\begin{tabular}{cc}
\hline \hline
\textbf{Original} & \textbf{Compressed} \\
\hline
\multicolumn{2}{c}{$z \in \mathbb{R}^{128}$} \\
\hline
\multicolumn{2}{c}{fc, 32 blocks, 128}  \\
\hline
\multicolumn{2}{c}{fc, 32 blocks, 256}  \\
\hline
fc, 32 blocks, 512  &  \\
\cline{1-1}
fc, 16 blocks, 1024 &  fc, 128 blocks, 784 \\
\cline{1-1}
fc, 16 blocks, 784 \\
\hline
\multicolumn{2}{c}{Tanh, reshape to $28\times28$} \\
\hline \hline
\end{tabular}
\renewcommand{\arraystretch}{1}
\caption{Fully connected RPGAN without nonlinearities and its compressed modification.}
\label{tab:fc_lin_gan}
\end{table}

\subsection{Linear map generator}
\label{sect:pure_linear}

As a surprising side effect of the RPGAN model, we discovered that decent generation quality can be achieved by the RPGAN generator \textbf{with no nonlinearities}, i.e., one can train the RPGAN generator with all instances consisting of linear transformations only. To demonstrate that, we take an RPGAN with the same ResNet-like generator architecture as in the experiments above. Then we replace all nonlinearities in the generator model by identity operations and train it on the CIFAR-10 dataset. The model demonstrates FID equal to $22.79$ that is competitive to the state-of-the-art generative models of comparable sizes. Note that this approach fails for a standard GAN generator that maps a Gaussian distribution to an images distribution. Indeed, that generator would be a linear operator from a latent space with a Gaussian distribution in the images domain.

\begin{figure}
    \centering
    \includegraphics[width=0.6\linewidth]{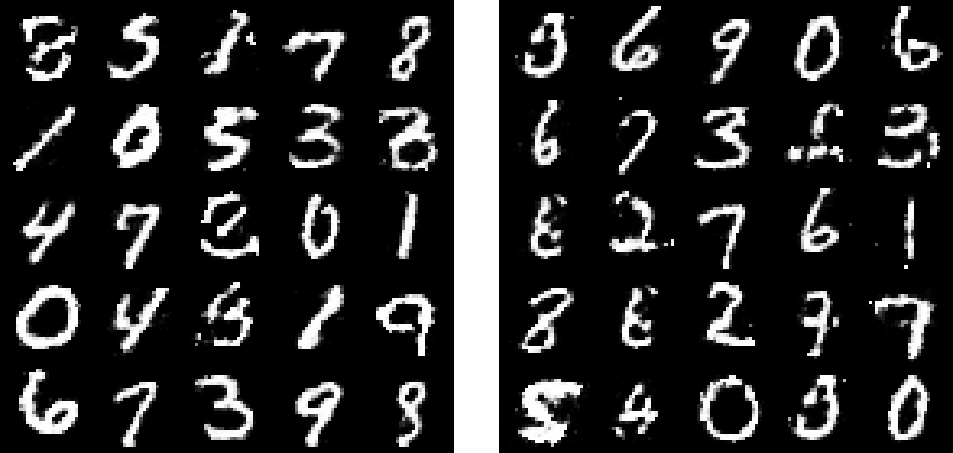}
    \vspace{-4mm}
    \caption{Digits generated by RPGAN without nonlinearities (\textit{left}) and by its $\times2.2$ faster compression (\textit{right}).}
    \vspace{-6mm}
    \label{fig:mnisct_compression}
\end{figure}

This purely linear generator architecture allows us to speed up the image generation process for fully-connected layers significantly. We group consequent buckets of fully-connected layers to form a new bucket. The instances in the new bucket are linear transformations that are products of the instances from the original buckets. To demonstrate this, we train a fully-connected generator network on the MNIST dataset, see \tab{fc_lin_gan}. Then we join the last three buckets into a single one. Namely, we form a new bucket by instances defined as the linear operators $B_{5k} \circ B_{4j} \circ B_{3i}$ where $i, j, k$ are random indices of instances from the buckets $B_3, B_4, B_5$ of the original generator. Thus instead of performing three multiplications of features vector from the second layer by matrices of the shapes $256\times512, 512\times1024, 1024\times784$, we perform a single multiplication by a $256\times784$ matrix. In our experiments, we achieved $\times2.2$ speed up. Note, however, that after the compression, the latent space cardinality can decrease if a small subset of tuples $\left(i,j,k\right)$ is used to populate the new bucket. Nevertheless, as random products of joining buckets are used, we expect that the generated images would be uniformly distributed in the space of images, produced by the uncompressed generator (see \fig{mnisct_compression} for visual comparison).

%% file: conclusion.tex
\section{Conclusion}
\label{sect:conclusion}

In this paper, we have proposed RPGAN --- the unsupervised technique to interpret and analyze GAN models. RPGAN is based on an alternative generator design that allows natural interpretation of different layers via using random routing as a source of stochasticity.  With experiments on several datasets, we provide evidence that different layers are responsible for the different factors of variation in generated images, which is consistent with findings from previous work. As a possible direction of future research, one can use the RPGAN analysis to construct efficient models, e.g., via identification of redundant parts of the generator for pruning or inference speedup.

%% file: appendix.tex
\onecolumn

\input{appendix/further_experiments}

%% file: appendix/further_experiments.tex
\section{Additional quantitative and qualitative results}
\label{sect:append:experiments}

\begin{figure}[h!]
    \centering
    \includegraphics[width=0.4\columnwidth]{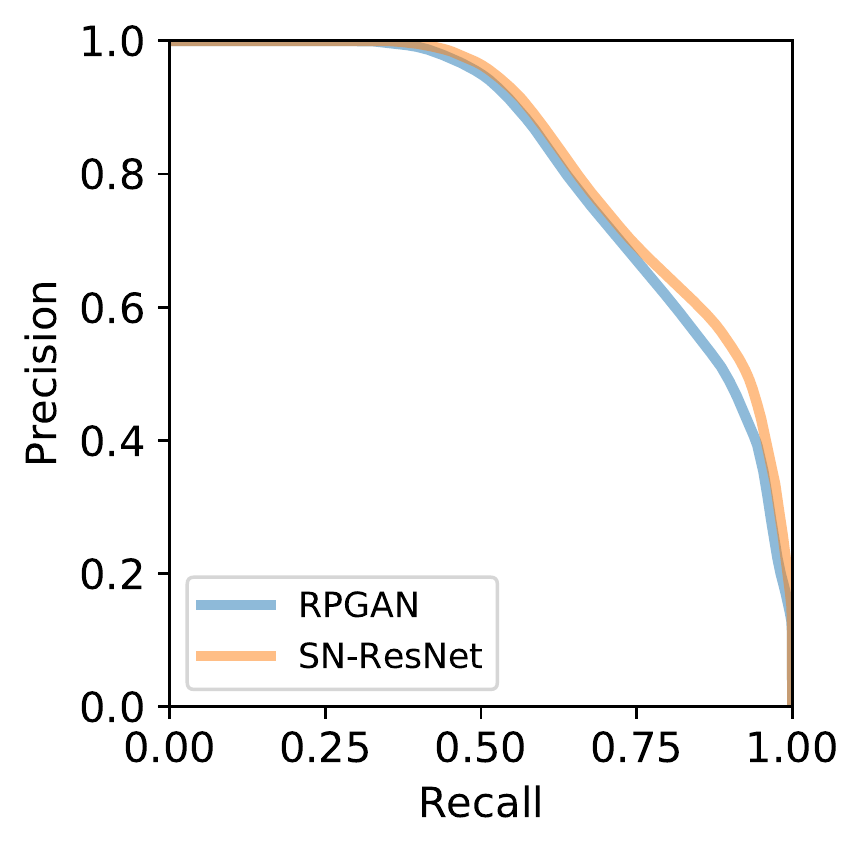}
    \caption{Precision-Recall curves for SN-ResNet GAN and RPGAN trained on CIFAR-10.}
    \label{fig:pr_cifan}
\end{figure}

\textbf{Generation quality.} To confirm that RPGAN provides the same generation quality as the standard GAN of the same backbone, we also compare them in terms of the recent precision-recall metrics \cite{precision_recall_distributions}, see \fig{pr_cifan}. It shows precision-recall curves for two models trained on the CIFAR-10 dataset and demonstrates almost equal generation quality. 

\begin{figure}
    \centering
    \includegraphics[width=0.99\columnwidth]{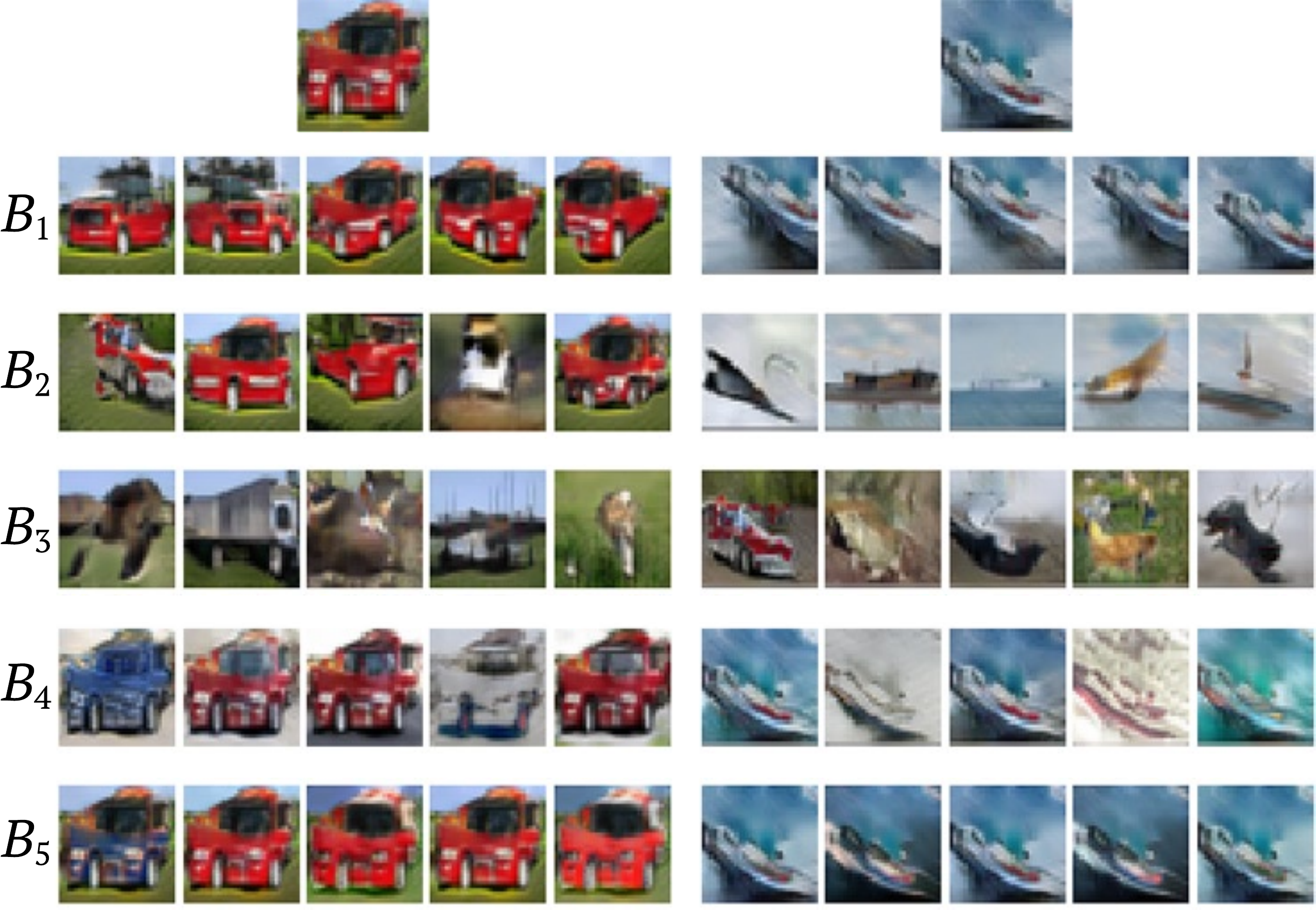}

    \vspace{0.5cm}

    \includegraphics[width=0.6\columnwidth]{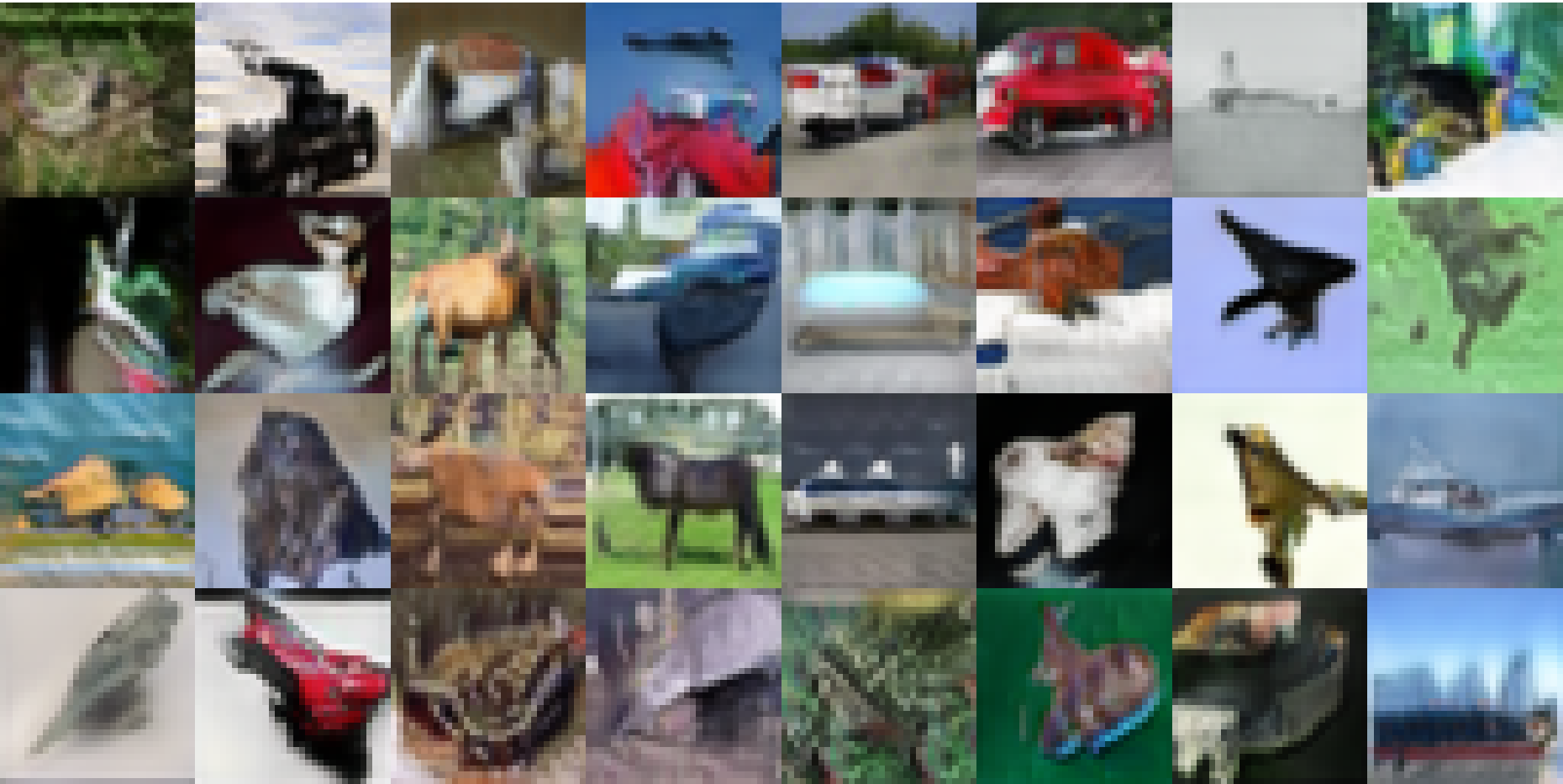}

    \caption{Images distributions for 5-bucket RPGAN and CIFAR-10 at resolution $32 \times 32$ (\textit{Top}). Random samples of the correspondent RPGAN (\textit{Bottom}).}
    \label{fig:sup_chart_cifar}
\end{figure}

\begin{figure}
    \centering
    \includegraphics[width=0.99\columnwidth]{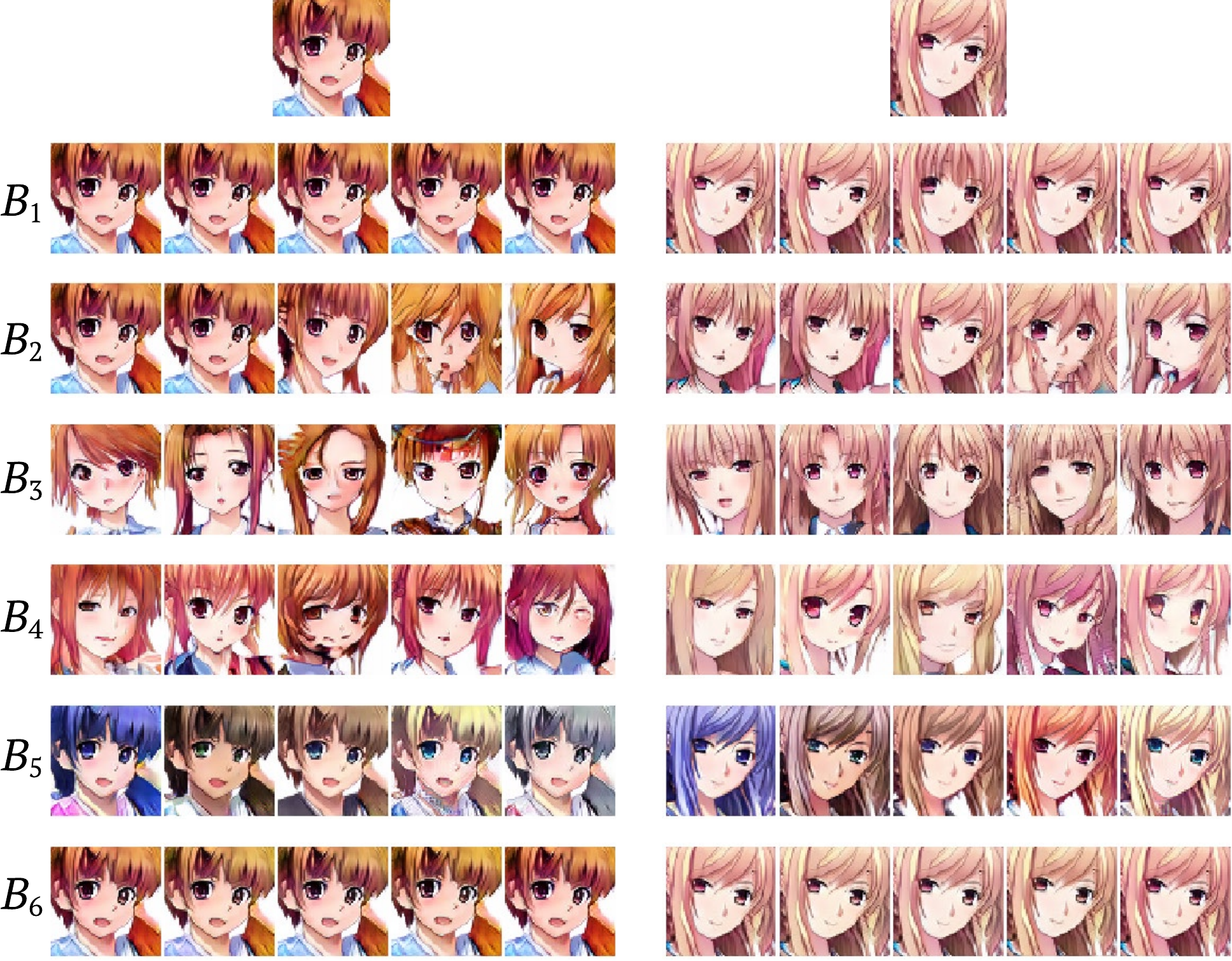}

    \vspace{0.5cm}

    \includegraphics[width=0.6\columnwidth]{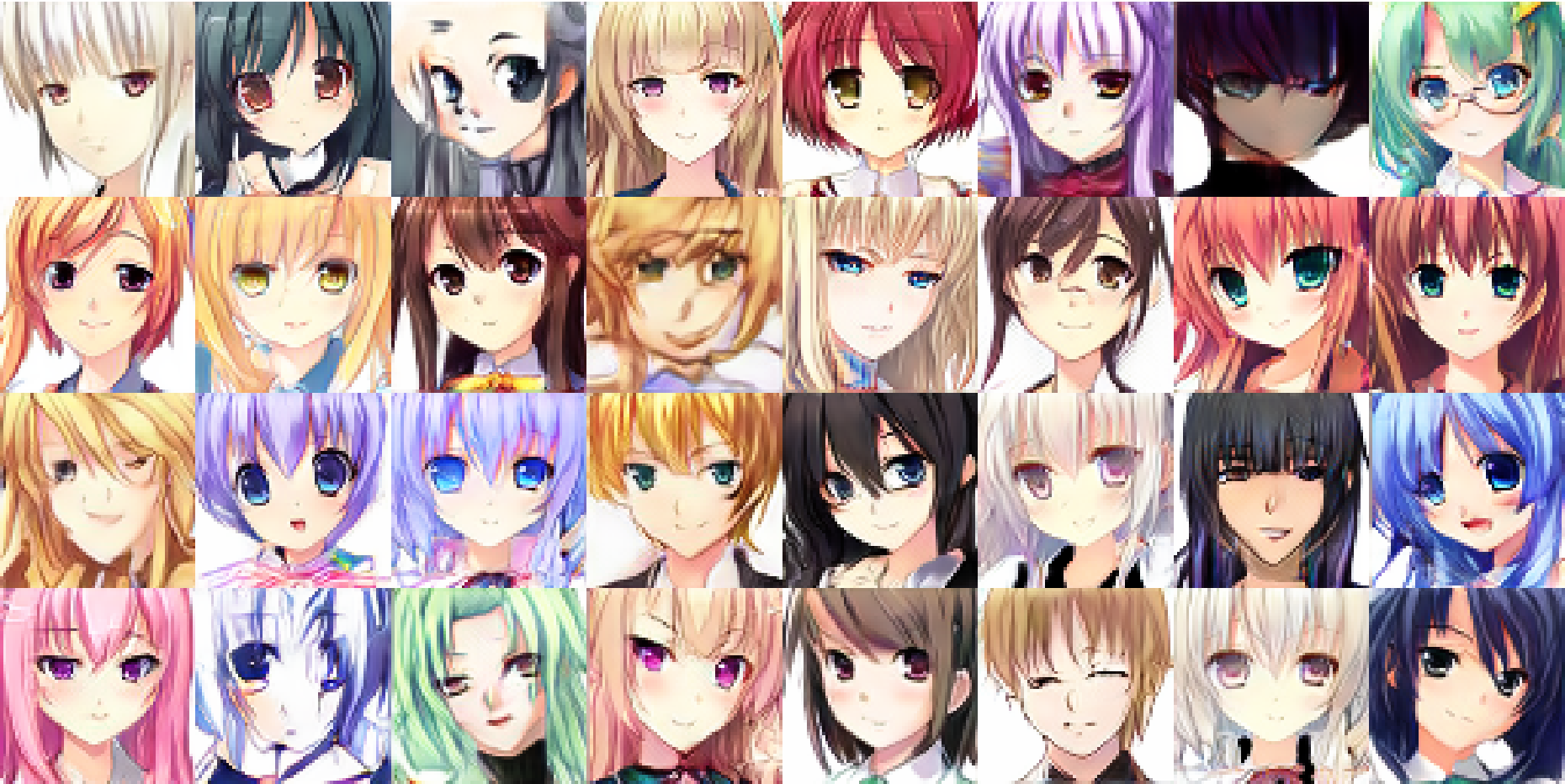}

    \caption{Images distributions for 6-bucket RPGAN and Anime Faces at resolution $64 \times 64$ (\textit{Top}). Random samples of the correspondent RPGAN (\textit{Bottom}).}
    \label{fig:sup_chart_anime}
\end{figure}

\begin{figure}
    \centering
    \includegraphics[width=0.99\columnwidth]{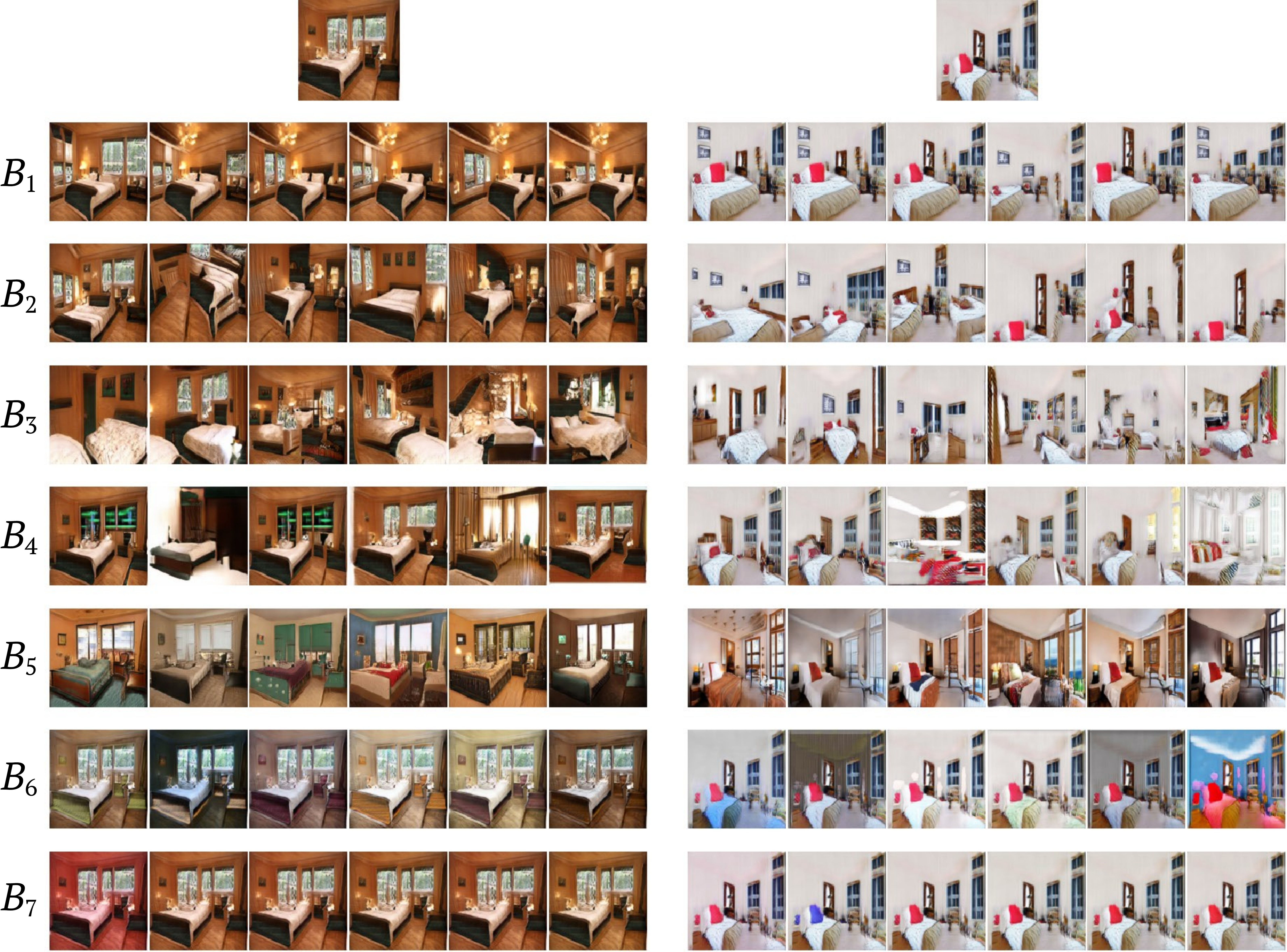}

    \vspace{0.5cm}

    \includegraphics[width=0.6\columnwidth]{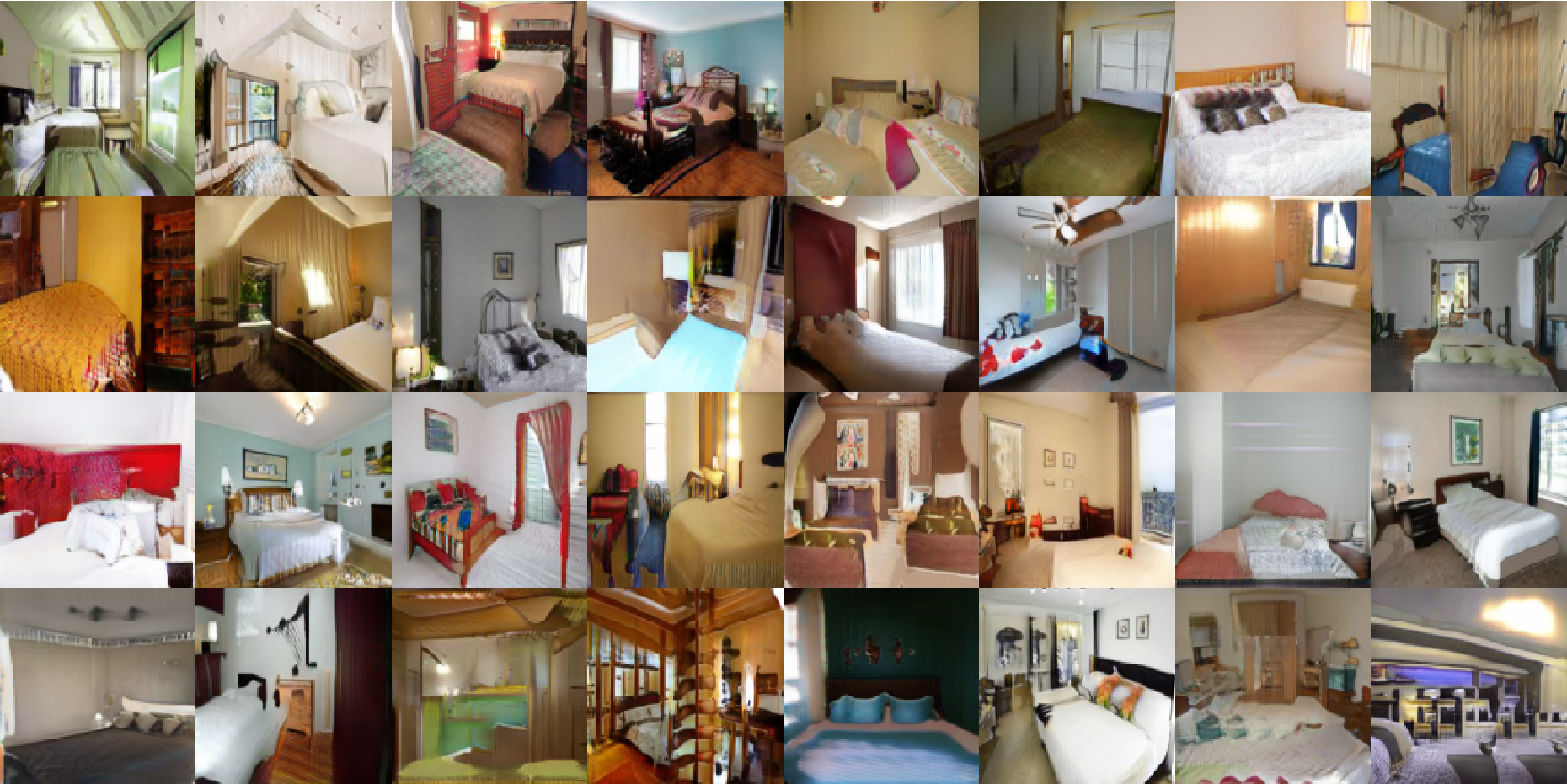}

    \caption{Images distributions for 7-bucket RPGAN and LSUN Bedroom at resolution $128 \times 128$ (\textit{Top}). Random samples of the correspondent RPGAN (\textit{Bottom}).}
    \label{fig:sup_chart_lsun}
\end{figure}

We also present several qualitative results of RPGAN generation on CIFAR-10 (\fig{sup_chart_cifar}) Anime Faces (\fig{sup_chart_anime}) and LSUN (\fig{sup_chart_lsun}). All the figures demonstrate that RPGAN does provide an understanding of responsibilities of different layers in generation.

\begin{figure}
    \centering
    \includegraphics[width=0.99\columnwidth]{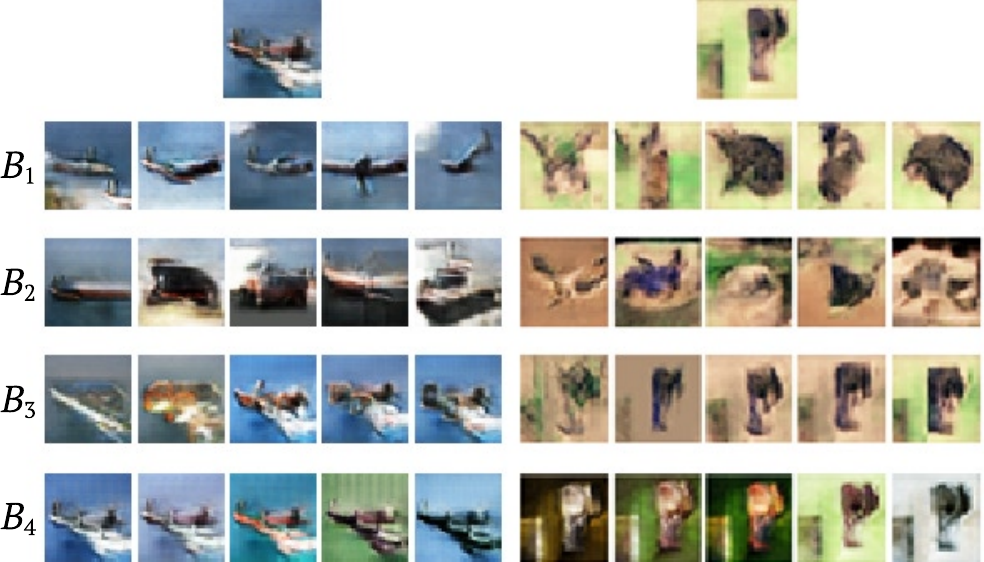}

    \caption{Images distributions for convolutional 4-bucket RPGAN and CIFAR-10 at resolution $32 \times 32$.}
    \label{fig:layers_variation_cifar10_wgan}
\end{figure}

\begin{figure}
    \centering
    \includegraphics[width=0.99\columnwidth]{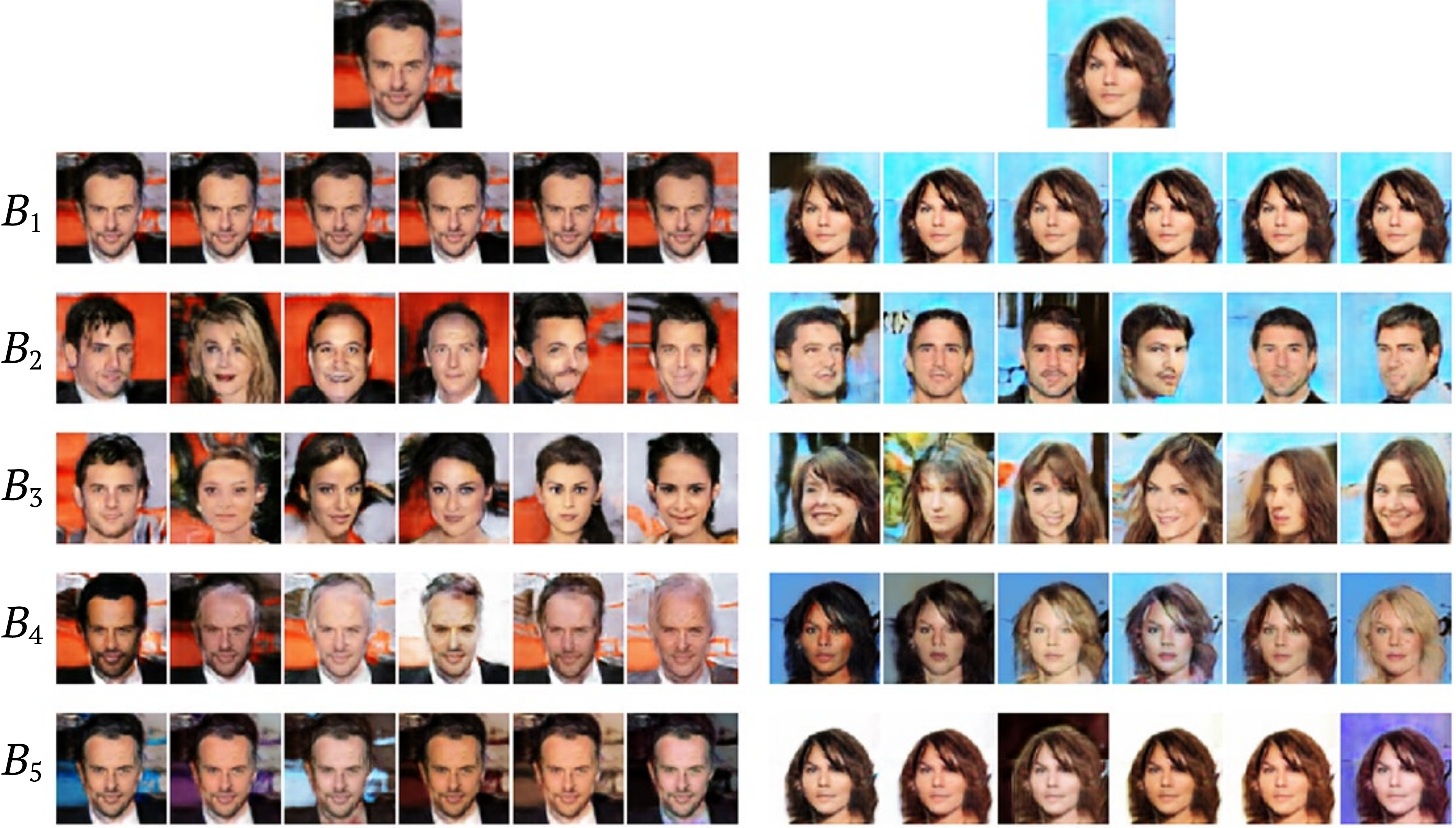}

    \caption{Images distributions for convolutional 5-bucket RPGAN and CelebA at resolution $64 \times 64$.}
    \label{fig:layers_variation_celeba_wgan}
\end{figure}

\begin{figure}
    \centering
    \includegraphics[width=0.99\columnwidth]{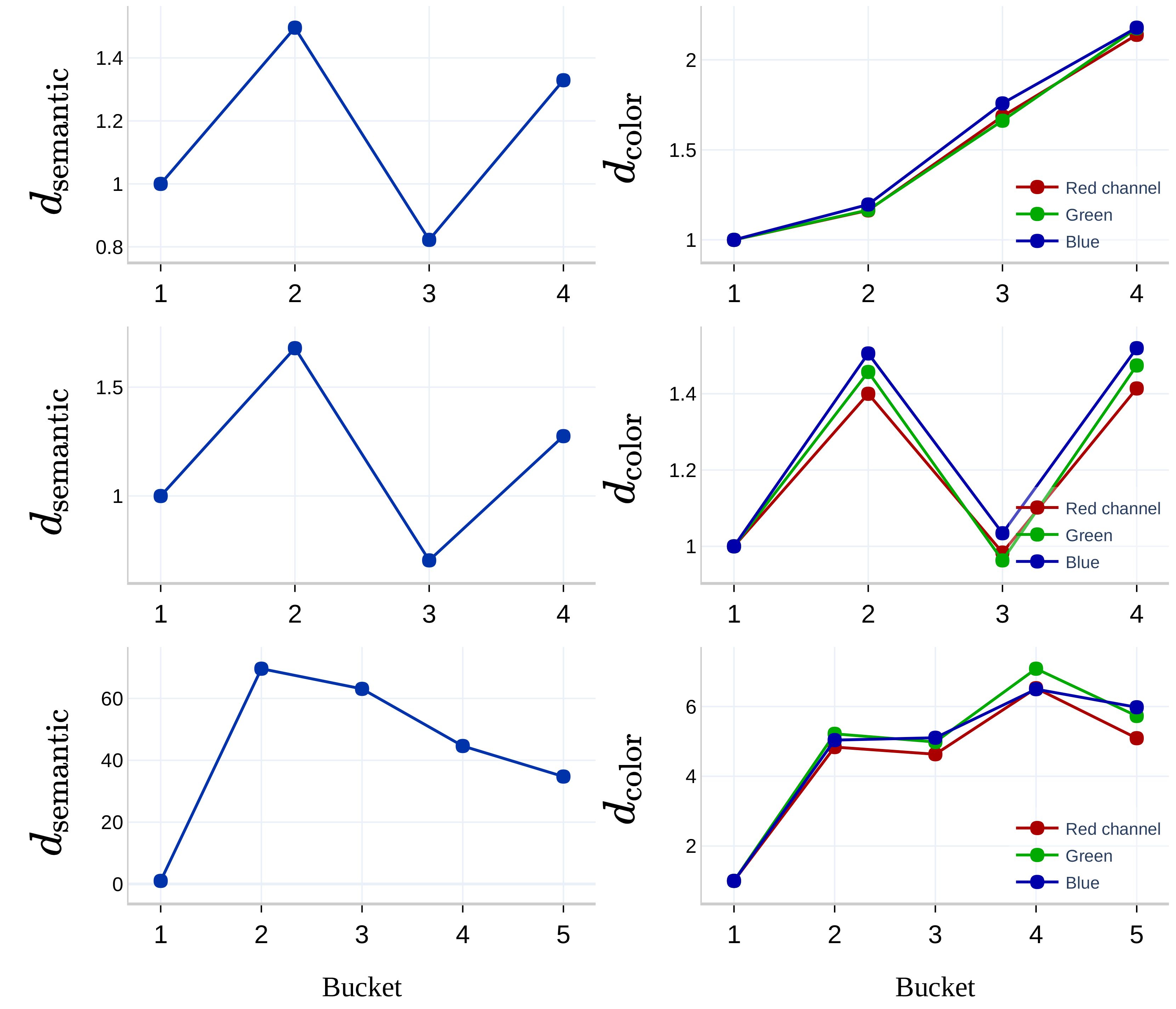}

    \caption{The quantitative evaluation of the extent $D_{l \to 1, d}$ to which different generator parts affect different factors of variation for (\textit{Top}) the four-bucket RPGAN and colored MNIST; (\textit{Middle}) the four-bucket RPGAN and CIFAR-10; (\textit{Bottom}) the five-bucket RPGAN and CelebA. In all cases, DCGAN-like generator architectures were used and each RPGAN bucket is associated with a convolutional layer.}
    \label{fig:layers_specification_sup}
\end{figure}

\textbf{Wasserstein GAN.} Here we show that RPGAN can be used for the analysis of different generator architectures and learning strategies. Namely, we present plots for DCGAN-like generators consisting of consequent convolutional layers without skip connections. All the models were trained with the same hyperparameters as described in Section $4$. Here we do not use spectral normalization and train generators as WGANs with weight penalty \cite{Gulrajani:2017:ITW:3295222.3295327}. On the \fig{layers_variation_cifar10_wgan} we show plots for a four-bucket generator trained on CIFAR-10. Additionally, we show plots for the five-bucket generator and CelebA-64x64 dataset on \fig{layers_variation_celeba_wgan}. See \fig{layers_specification_sup} for the quantitative evaluation for these two GANs and the RPGAN trained on colored MNIST dataset introduced in Section 4.2.

\section{RPGAN inversion for image editing}

\begin{figure}
    \centering
    \includegraphics[width=0.99\columnwidth]{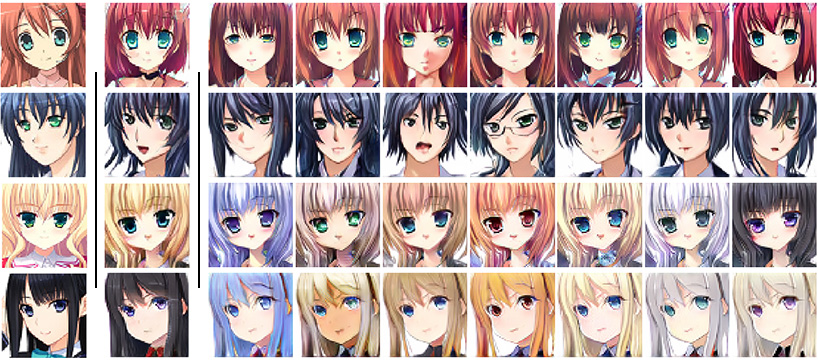}

    \caption{Reconstructions of the real data images samples. \textit{First column}: real data sample; \textit{Second column}: its reconstruction with the invertor $E$; \textit{Other images in lines}: the images generated by the RPGAN by replacing a fixed bucket reconstructed index. For the first two images, we modify the bucket responsible for semantics, while for the last two we modify the bucket responsible for coloring.}
    \label{fig:anime_rec}
\end{figure}

The discrete nature of the RPGAN latent space implies a simple procedure for RPGAN inversion, which can be useful for image editing. Let we have an RPGAN with buckets $B_1, \dots, B_n$ with a number of instances $m_1, \dots, m_n$ respectively. Our goal then is to obtain an \textit{encoder} $E: I \to \left<m_1\right> \times \cdots \times \left<m_n\right>$ from the image space to a cartesian product of indices. For $E$ we take $n$ independent Network in Network \cite{nin} classification models, where each model is trained to predict an index of the instance that was used for a particular image generation. We train these models for six-bucket RPGAN and the Anime Faces dataset on generated images. The accuracy values of classification models are provided in \tab{rec_acc}. As one can see, for the first two buckets, it is difficult to predict the corresponding instance indices. Meanwhile, instance indices from the latter buckets can be predicted almost perfectly. Given encoder, image editing can be performed as follows. For a real image, we obtain its \textit{reconstruction} in $\left<m_1\right> \times \cdots \times \left<m_n\right>$ with $E$ and then ``tweak'' the instances from different buckets to perform semantic manipulations. Examples of reconstructions and editing are presented on \fig{anime_rec}.

\begin{table}[!h]
\addtolength{\tabcolsep}{-4pt}
\centering
    \begin{tabular}{|c|c|c|c|c|c|c|c|c|c|c|}
        \hline
        Bucket & $1$ & $2$ & $3$ & $4$ & $5$ & $6$\\
        \hline
        Accuracy & $0.05$ & $0.21$ & $1.0$ & $0.99$ & $1.0$ & $0.69$ \\
        \hline
    \end{tabular}
\caption{Instances inversion accuracies for different buckets. All buckets are consisted of 20 instances. Evaluated on $1280$ generated images.}
\label{tab:rec_acc}
\end{table}

